\documentclass[sigconf,nonacm]{acmart}

\usepackage{booktabs}
\usepackage{graphicx}
\usepackage{multirow}
\usepackage{amsmath}
\usepackage{xcolor}
\usepackage{subcaption}
\usepackage{tikz}
\usepackage{algorithm}
\usepackage{algorithmic}
\usetikzlibrary{positioning, arrows.meta, calc, fit, backgrounds, shapes.geometric}

\newcommand{\feat}[1]{\texttt{\small #1}}
\graphicspath{{../data/results/gan_figures/}{../data/results/replay_figures/}{../data/results/optimization_figures/}{./}}

\renewcommand\footnotetextcopyrightpermission[1]{}
\begin{document}

\title{What Does It Take to Detect an AI Agent? Minimal Feature Sets for Behavioral Detection under Browser Automation}
\titlenote{Accepted at the North East AI Agents Day 2026 workshop, Jane Street, New York City. \url{https://ne-agents-day.github.io/}}

\author{Vishisht Choudhary}
\email{vishisht.choudhary@tum.de}
\affiliation{%
  \institution{Technical University of Munich}
  \city{Munich}
  \country{Germany}
}

\author{Lukas Schmidt}
\email{lukas.s.schmidt@tum.de}
\affiliation{%
  \institution{Technical University of Munich}
  \city{Munich}
  \country{Germany}
}

\author{Anne Zoë Kenntner}
\email{annezoe.kenntner@tum.de}
\affiliation{%
  \institution{Technical University of Munich}
  \city{Munich}
  \country{Germany}
}

\author{Feras Skhab}
\email{feras.skhab@tum.de}
\affiliation{%
  \institution{Technical University of Munich}
  \city{Munich}
  \country{Germany}
}

\author{Michel Osswald}
\email{michel@kontext.security}
\affiliation{%
  \institution{Kontext}
  \city{Munich}
  \country{Germany}
}

\author{Jens Ernstberger}
\email{jens@kontext.security}
\affiliation{%
  \institution{Kontext}
  \city{Munich}
  \country{Germany}
}

\begin{abstract}
    Bot detectors deployed at scale treat traffic as binary: human or bot.
    This assumption breaks in a world where AI agents browse the web through
    browser automation, a traffic class that is neither and that binary
    classifiers structurally cannot represent.
    
    We present a three-class detection framework that correctly distinguishes
    humans, bots, and AI agents, and show that the binary-vs-agent confusion
    is architectural: a binary human-vs-bot detector \emph{misroutes} agent
    sessions because the label space lacks an agent class. On our controlled
    benchmark, an MLP binary classifier misclassifies 39.1\% of real AI
    agents as human and a SAINT binary transformer misclassifies 34.5\%;
    adding an explicit agent class yields per-class agent F1\,=\,1.000 in
    every one of 30 runs (3 model families $\times$ 10 seeds), eliminating
    the gap entirely.

    To measure how evasion-resistant this detection is, we construct a
    five-level evasion ladder that spans passive observation,
    GAN-generated trajectories, and replay of real human cursor data
    ($n = 2{,}299$ archived evasion sessions: 1{,}025 unmodified agent +
    150 rule-based + 300 GAN + 300 GAN+ + 524 human replay). Across 10
    seeds and 3 model families we observe zero agent misses in 22{,}990
    per-seed predictions: 100\% agent recall at every level, with macro-F1
    95\% CIs of $[0.986, 0.994]$ (SAINT), $[0.993, 0.997]$ (RF), and
    $[0.993, 0.998]$ (GBM). The discriminative signal is a
    browser-automation artifact, not evidence of agent reasoning.
    Playwright does not emit the raw pointer-move and wheel-delta streams
    that a physical input device produces, and this absence signature
    (no raw-event teleports before clicks, no wheel deltas, empty
    pre-click mousemove tracks) survives trajectory manipulation.

    Exhaustive search over all $\binom{17}{k}$ feature subsets
    ($k = 1\ldots5$; 9{,}401 GBMs trained, with full per-class precision
    and recall) shows that a handful of event-stream features recovers
    strong agent detection. Two behavioral features
    (\feat{mouse\_event\_rate}, \feat{teleport\_click\_ratio}) jointly
    give 100\% observed agent recall at every evasion level, with held-out
    test macro-F1 of 0.926 and agent precision of 0.994; five features
    lift macro-F1 to 0.991. Backward elimination finds a different
    two-feature set (\{\feat{mouse\_event\_rate},
    \feat{click\_duration\_std}\}) that also preserves 100\% agent recall
    at unit precision: the signal is redundantly encoded, and removing
    \feat{teleport\_click\_ratio} entirely leaves agent detection at
    100\%. The single-feature regime is degenerate. The one indicator that
    flags every agent (\feat{cursor\_path\_linearity}) does so by
    collapsing the classifier to always predict ``agent'' (agent precision
    0.33, macro-F1 0.17), a failure mode our per-level diagnostics surface
    automatically. \textbf{Two features robustly isolate agents; five
    separate all three traffic classes at macro-F1 $\geq 0.99$.}\end{abstract}


\maketitle

\section{Introduction}
\label{sec:introduction}

The way we interact with the internet is changing. For decades, web traffic fell into two categories: \emph{humans} who browse with intent and motor variability, and \emph{bots}: custom scripts, scrapers, and automation frameworks that interact programmatically~\cite{iliou2021detection,jonker2019fingerprint}. The security industry built an entire detection ecosystem around this binary distinction: CAPTCHAs~\cite{vonahn2008recaptcha}, behavioral biometrics~\cite{pusara2004mouse,ahmed2007mouse}, and commercial bot management platforms all ask the same question: \emph{human or not?}

A third actor has now entered the web. LLM-driven agents (systems like Anthropic's Computer Use~\cite{anthropic2024computeruse}, OpenAI's Operator~\cite{openai2025operator}, and research frameworks such as WebArena~\cite{zhou2024webarena}, BrowserGym~\cite{drouin2024browsergym}, and Mind2Web~\cite{deng2024mind2web}) autonomously browse the web, reading pages, reasoning about content, and executing multi-step tasks through real browsers. Unlike traditional bots, agents generate trusted DOM events, adapt their behavior to page content, and pursue goals that require understanding. The distinction between who and what is on the internet is no longer binary.

Industry has begun to treat agents as a distinct traffic class. Cloudflare's Signed Agents protocol~\cite{cloudflare2024signedagents}, for instance, lets agents cryptographically self-identify so platforms can route, rate-limit, and serve each traffic class appropriately. But when agents do not declare themselves, platforms fall back on behavioral detection, which today remains binary: human or bot.

This raises a fundamental question: \textbf{are agents just another kind of bot?} If so, existing binary detectors should catch them. We show that an MLP binary classifier lets 39.1\% of AI agents through as ``human,'' a SAINT tabular transformer~\cite{somepalli2021saint} lets 34.5\% through, and a random forest lets 30.0\% through. Only XGBoost catches every agent, and it does so via \emph{coincidental} features (missing scroll events, absent raw mousemove streams before clicks, empty wheel-delta series) rather than agent-specific signals, so its detection is brittle. The gap is structural, not a matter of modeling capacity: a binary label space cannot represent a three-class world. Adding a third ``agent'' class eliminates the gap entirely (F1\,=\,1.000). We then show that the discriminative signal arises from how browser automation APIs generate events (programmatic DOM calls rather than hardware input), not from LLM reasoning cycles. That leaves a practical question: deploying 16+ behavioral features is expensive. How many do you \emph{actually need}?

We study this question in the setting of API-driven browser automation (Playwright), where agents interact with web pages through programmatic interfaces that generate trusted DOM events. We focus on Playwright because it is the dominant browser-driver substrate of the current agent ecosystem: Browser Use~\cite{browseruse}, Stagehand (Browserbase)~\cite{stagehand}, Skyvern~\cite{skyvern}, and most LangChain/MCP browser tools dispatch their actions through Playwright or the near-identical Puppeteer/CDP path, so its DOM-event signature is what deployed web agents typically emit in practice. Our dataset comprises more than 14{,}000 real human sessions (CaptchaSolve30k), 5{,}000 synthetic bot samples across three archetypes, and 1{,}025 real agent sessions recorded from Claude driving Chrome via Playwright across ten web tasks grouped into five interaction types. Our contributions:

\begin{enumerate}
    \item We show that binary human--bot detectors miss a structural fraction of AI agents (MLP 39.1\%, SAINT 34.5\%, random forest 30.0\%) and that this gap is architectural: a three-class formulation eliminates it entirely (\S\ref{sec:gap}).
    \item We show that the discriminative signal is an API execution artifact rather than an agent cognition signal: detection survives even when agents replay real human cursor trajectories, because automation APIs still dispatch actions via programmatic DOM calls (\S\ref{sec:ladder}, \S\ref{sec:discussion}).
    \item We evaluate robustness under five progressively stronger evasion strategies, from no evasion through GAN-generated trajectories to replay of real human cursor data, and observe 100\% agent detection at all levels (\S\ref{sec:ladder}).
    \item We identify \textbf{minimal feature subsets for practical deployment} through exhaustive search over $\binom{17}{k}$ subsets ($k$\,=\,1\ldots5; 9{,}401 GBMs), forward selection, and backward elimination (\S\ref{sec:optimization}):
    \begin{itemize}
        \item \textbf{2 behavioral features} suffice for 100\% agent recall at every evasion level with agent precision $\geq$\,0.99: \{\feat{mouse\_event\_rate}, \feat{teleport\_click\_ratio}\} from exhaustive search, or \{\feat{mouse\_event\_rate}, \feat{click\_duration\_std}\} from backward elimination (held-out test macro-F1 0.926 and 0.904, respectively).
        \item The $k$\,=\,1 regime is degenerate: the one solo-``perfect'' indicator (\feat{cursor\_path\_linearity}) flags every agent only by collapsing the predictor to always output ``agent'' (agent precision = 0.33, macro-F1 = 0.17).
        \item \textbf{5 features} reach macro-F1 $\geq$\,0.99 for full three-class discrimination; 17 features reach 0.995.
        \item The signal is redundantly encoded: removing \feat{teleport\_click\_ratio} entirely leaves agent recall at 100\% because \feat{mouse\_event\_rate}, \feat{click\_duration\_std}, and the missingness-indicator channels absorb the signature (\S\ref{sec:optimization}).
    \end{itemize}
\end{enumerate}

To our knowledge, this is the first work to study LLM-driven browser agents as a detection category distinct from traditional bots, and the first to perform exhaustive feature subset analysis for behavioral agent detection. We detail our dataset construction (\S\ref{sec:dataset}), feature engineering (\S\ref{sec:features}), model architecture (\S\ref{sec:model}), and evasion methodology (Appendix~\ref{sec:evasion}) to enable full reproducibility.

\section{Background \& Related Work}
\label{sec:related}

The evolution of non-human web traffic mirrors the evolution of web interaction itself. Each generation of automated actors prompted a new generation of defenses, and each defense assumed that the actor taxonomy of its day was complete. We trace this progression to motivate why agents require a new detection category.

\subsection{The Two-Actor Web: Humans and Bots}

For most of the web's history, non-human traffic meant \emph{bots}: scripts, scrapers, and automation frameworks that interact with websites programmatically. Early defenses relied on CAPTCHAs~\cite{vonahn2008recaptcha}, but as bots grew more sophisticated, the field shifted to behavioral biometrics: analyzing mouse dynamics~\cite{pusara2004mouse,ahmed2007mouse}, keystroke patterns~\cite{mondal2017continuous}, and browser fingerprints~\cite{jonker2019fingerprint} to detect non-human behavior without explicit challenges. Iliou et~al.~\cite{iliou2021detection} combined web server logs with mouse biometrics to detect advanced bots. Acien et~al.~\cite{acien2021becaptcha} proposed BeCAPTCHA-Mouse, using GAN-generated synthetic mouse trajectories to harden bot detection via data augmentation. Chu et~al.~\cite{chu2018bot} showed that mouse dynamics carry discriminative signals even in online gaming. Commercial platforms (reCAPTCHA, hCaptcha, Akamai, Cloudflare, HUMAN) deploy these techniques at scale.

All of these systems share a common assumption: traffic is either human or bot. The entire detection ecosystem, from feature design through model training to deployment thresholds, is built around this binary distinction. For two decades this worked, because the taxonomy was complete.

\subsection{The Third Actor: LLM-Driven Agents}

The emergence of LLM-driven web agents breaks this taxonomy. ReAct~\cite{yao2023react} demonstrated that language models can synergize reasoning and acting; WebGPT~\cite{nakano2022webgpt} showed browser-assisted question-answering with human feedback. These foundational ideas have spawned an ecosystem of agent frameworks (WebArena~\cite{zhou2024webarena}, BrowserGym~\cite{drouin2024browsergym}, Mind2Web~\cite{deng2024mind2web}, SeeAct~\cite{zheng2024seeact}) and commercial products: Anthropic's Computer Use~\cite{anthropic2024computeruse} and OpenAI's Operator~\cite{openai2025operator}.

What makes agents categorically different from bots? Bots follow predetermined scripts; agents \emph{reason} about page content and adapt. Bots produce mechanical, repetitive interaction patterns; agents exhibit a characteristic \emph{pause-think-act} cycle driven by LLM inference. Most critically, agents operate through real browsers via automation APIs (Playwright, Puppeteer~\cite{puppeteer}, Selenium~\cite{selenium}), generating trusted DOM events that look nothing like the headless requests of traditional scrapers. This places agents in an uncanny middle ground for the most nuanced binary detectors: MLP fails on 39.1\% of agents and SAINT fails on 34.5\%, while simpler models catch them via coincidental features (\S\ref{sec:gap}). Either way, binary detectors cannot \emph{route} agents as a distinct class, and their event-stream structure remains fundamentally non-human.

Our 39.1\% (MLP) / 34.5\% (SAINT) evasion result quantifies this gap. When the most nuanced binary detectors encounter agents, they classify roughly one in three as human, not because the detectors are weak but because agents genuinely do not fit the behavioral profile of bots. They are a third category that binary detectors were never designed to handle \emph{via agent-specific signals}: XGBoost in our benchmark catches 100\% of agents, but via coincidental features such as missing scroll events and empty mousemove streams, and such detection is brittle.

\subsection{Detection Methods: From Trees to Transformers}

Given a three-class detection problem, classifier choice matters. Our problem is fundamentally tabular: given a vector of behavioral features, classify traffic as human, bot, or agent. Grinsztajn et~al.~\cite{grinsztajn2022tree} showed that tree-based models (XGBoost~\cite{chen2016xgboost}, Random Forest) still outperform deep learning on many tabular benchmarks, particularly with heterogeneous features and small datasets. Yet transformer-based architectures have closed this gap: SAINT~\cite{somepalli2021saint} uses per-feature embedding with row attention and contrastive pre-training; TabTransformer~\cite{huang2020tabtransformer} applies self-attention to categorical features; FT-Transformer~\cite{gorishniy2021revisiting} treats all features uniformly through per-feature projections; TabNet~\cite{arik2021tabnet} uses sequential attention for feature selection; and NODE~\cite{popov2020neural} combines oblivious decision trees with gradient-based optimization.

We choose SAINT because it is among the most vulnerable to agents in the binary setting (34.5\% evasion, second only to MLP's 39.1\%) and because it learns the most nuanced decision boundaries between humans and bots, which means agents fall closest to the human side. If we can fix SAINT's blind spot with a three-class formulation, the result is more meaningful than fixing a model that already catches agents by coincidence. We adapt SAINT with denoising pre-training and a multi-class output head (\S\ref{sec:model}).

\subsection{Adversarial Evasion: How Hard Is It to Hide?}

If agents are detectable, the natural follow-up is: can they evade? Adversarial machine learning has shown that classifiers can be fooled by crafted inputs~\cite{szegedy2014intriguing,goodfellow2015adversarial}, and optimization-based attacks can defeat even defensive distillation~\cite{carlini2017adversarial}. In security domains, Grosse et~al.~\cite{grosse2017adversarial} demonstrated adversarial evasion of malware classifiers.

For behavioral biometrics, evasion means generating human-like interaction patterns. GANs~\cite{goodfellow2014generative} have been used to synthesize realistic mouse trajectories~\cite{acien2021becaptcha}, and replay attacks, where recorded human behavior is replayed verbatim, represent the theoretical ceiling for trajectory-level evasion. Our five-level evasion ladder (Appendix~\ref{sec:evasion}) systematically escalates from no evasion through GAN-generated trajectories to direct replay of real human cursor data, testing whether detection relies on surface-level patterns (matchable) or fundamental event-stream structure (not matchable).

\subsection{Feature Selection: How Little Can You Get Away With?}

Even with robust detection, deploying 16+ features in production is expensive: each feature requires client-side instrumentation, server-side extraction, and ongoing maintenance. Feature selection, choosing the minimal subset that preserves performance, is therefore critical for deployment. The literature distinguishes three paradigms~\cite{guyon2003introduction}: \emph{filter} methods rank features by statistical tests (e.g., mutual information~\cite{peng2005mrmr}); \emph{wrapper} methods evaluate subsets using the target classifier; and \emph{embedded} methods perform selection during training (e.g., LASSO~\cite{tibshirani1996lasso}).

For $n$ features, exhaustive wrapper search requires evaluating $2^n - 1$ subsets, which is feasible only for small $n$. With 17 continuous features (16 base + \feat{scroll\_interval\_variance} added after the resampler bug fix described in the Methodology), we enumerate all $\sum_{k=1}^{5}\binom{17}{k} = 9{,}401$ subsets for $k$\,=\,1\ldots5, complemented by greedy forward selection and backward elimination. This gives practitioners a concrete answer: not just \emph{can} you detect agents, but \emph{what is the least you need to deploy}?

\section{Dataset Construction}
\label{sec:dataset}

\begin{figure}[t]
\centering
\resizebox{\columnwidth}{!}{%
\begin{tikzpicture}[
    node distance=0.4cm and 0.6cm,
    source/.style={draw, rounded corners, minimum height=0.6cm, minimum width=1.8cm, font=\footnotesize, fill=#1, align=center},
    source/.default=blue!8,
    proc/.style={draw, rounded corners, minimum height=0.5cm, font=\footnotesize, fill=gray!10, align=center},
    arrow/.style={-Stealth, semithick},
]
\node[source=orange!15] (human) {CaptchaSolve30k\\ \scriptsize 14k+ sessions};
\node[source=red!15, below=0.5cm of human] (bot) {Synthetic Gen.\\ \scriptsize 3 archetypes};
\node[source=blue!15, below=0.5cm of bot] (agent) {Claude + Playwright\\ \scriptsize 1{,}025 sessions};

\node[proc, right=0.8cm of bot] (extract) {Feature\\ Extraction};
\node[proc, right=0.6cm of extract] (balance) {Balance};
\node[proc, right=0.6cm of balance] (unified) {\textbf{Balanced}\\ \scriptsize 3{,}075$\times$28};

\node[font=\scriptsize, left=0.05cm of human, text=orange!70!black] {0};
\node[font=\scriptsize, left=0.05cm of bot, text=red!70!black] {1};
\node[font=\scriptsize, left=0.05cm of agent, text=blue!70!black] {2};

\draw[arrow] (human.east) -- (extract);
\draw[arrow] (bot.east) -- (extract);
\draw[arrow] (agent.east) -- (extract);
\draw[arrow] (extract) -- (balance);
\draw[arrow] (balance) -- (unified);
\end{tikzpicture}%
}
\caption{Dataset construction pipeline. Human behavioral data from CaptchaSolve30k, synthetic bot traffic from a three-archetype generator, and real AI agent sessions are merged, feature-extracted, and balanced to produce the unified training set.}
\label{fig:dataset_overview}
\end{figure}
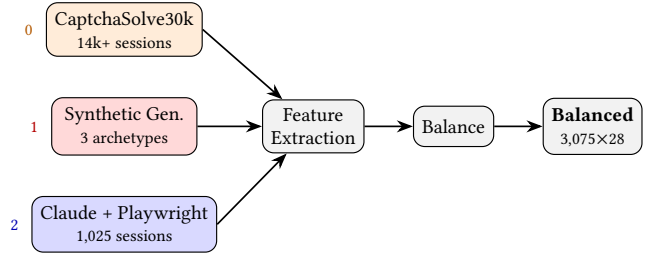

We construct a three-class dataset comprising human behavioral data, synthetic bot traffic, and real AI agent sessions. Figure~\ref{fig:dataset_overview} summarizes the data pipeline.

\paragraph{Scope of claims.} The bot class is sampled from literature-grounded distributions, so \emph{three-class metrics} (macro-F1, the minimal feature set for three-class discrimination) should be interpreted relative to this controlled bot model. \emph{Agent detection}, however, is evaluated on real data for both the human and agent classes: the 39.1\%/34.5\% binary gap and the 100\% three-class agent detection are bot-simulation-independent.

\subsection{Human Behavioral Data}

We source human interaction data from CaptchaSolve30k~\cite{captchasolve30k}, a public dataset of mouse dynamics recorded during real CAPTCHA-solving tasks on the HuggingFace Hub. The dataset contains mouse trajectories from over 14{,}000 sessions, each capturing cursor movements, clicks, and timing as users solve visual CAPTCHAs. Event parsing, click-edge reconstruction, trajectory segmentation, and 60\,Hz resampling details are given in Appendix~\ref{app:data}.

\subsection{Synthetic Bot Generation}
\label{sec:botgen}

Real bot traffic is difficult to obtain at scale with ground-truth labels. We generate synthetic bot data using a three-archetype model that reflects the real-world spectrum of bot sophistication~\cite{iliou2021detection}:

\begin{itemize}
    \item \textbf{Headless bots} (35\% of samples) model browser automation without a visible browser window~\cite{puppeteer,selenium}. These bots produce no mouse, scroll, or keyboard events; 85\% expose the \texttt{navigator.webdriver} flag; and 70\% report zero viewport dimensions.
    \item \textbf{Scripted bots} (35\%) model Selenium/Puppeteer-driven automation with basic mouse simulation. They produce low trajectory entropy ($\mathcal{N}(0.5, 0.3)$), high movement efficiency ($\mathcal{N}(0.92, 0.05)$), fast uniform clicks ($\mathcal{N}(50, 15)$\,ms mean duration), and high teleport-click ratios ($\mathcal{N}(0.55, 0.2)$).
    \item \textbf{Sophisticated bots} (30\%) model behavior-mimicking automation with deliberate humanization. Their distributions intentionally overlap with human behavior: trajectory entropy $\mathcal{N}(1.8, 0.6)$, click duration $\mathcal{N}(85, 25)$\,ms, and inter-event entropy $\mathcal{N}(2.5, 0.7)$.
\end{itemize}

All features are drawn from clipped normal distributions to ensure physical plausibility (e.g., non-negative durations, efficiencies in $[0,1]$). Table~\ref{tab:distributions} summarizes the distribution parameters for all archetypes alongside the human baseline.

\begin{table}[t]
\centering
\caption{Distribution parameters for synthetic bot generation. Each feature is drawn from $\mathcal{N}(\mu, \sigma)$ clipped to the specified range. Human parameters are estimated from CaptchaSolve30k.}
\label{tab:distributions}
\footnotesize
\begin{tabular}{lcccc}
\toprule
Feature & Human & Headless & Scripted & Sophisticated \\
\midrule
traj\_entropy     & $\mathcal{N}(2.8, 0.8)$ & --- & $\mathcal{N}(0.5, 0.3)$ & $\mathcal{N}(1.8, 0.6)$ \\
move\_efficiency  & $\mathcal{N}(0.55, 0.15)$ & --- & $\mathcal{N}(0.92, 0.05)$ & $\mathcal{N}(0.72, 0.12)$ \\
click\_dur (ms)   & $\mathcal{N}(110, 30)$ & --- & $\mathcal{N}(50, 15)$ & $\mathcal{N}(85, 25)$ \\
typing\_spd (ms)  & $\mathcal{N}(180, 60)$ & --- & $\mathcal{N}(50, 20)$ & $\mathcal{N}(120, 40)$ \\
inter\_evt\_H     & $\mathcal{N}(3.5, 0.8)$ & --- & $\mathcal{N}(1.0, 0.5)$ & $\mathcal{N}(2.5, 0.7)$ \\
evt\_rate (Hz)    & $\mathcal{N}(15, 7)$ & --- & $\mathcal{N}(2, 1.5)$ & $\mathcal{N}(8, 4)$ \\
teleport\_ratio   & $\mathcal{N}(0.05, 0.06)$ & --- & $\mathcal{N}(0.55, 0.2)$ & $\mathcal{N}(0.18, 0.1)$ \\
\bottomrule
\multicolumn{5}{l}{\scriptsize ``---'' indicates no events generated (headless bots have no interaction).}
\end{tabular}
\end{table}

We generate 5{,}000 synthetic bot samples balanced across the three archetypes. The sophisticated archetype is deliberately designed to overlap with human distributions, making the bot-vs-human boundary the hardest classification challenge, an important design choice: it ensures our detector cannot rely on trivially separable features.


\subsection{Agent Data Collection}

We collect real AI agent behavioral data using Claude~\cite{anthropic2024computeruse} driving a Chrome browser via the Model Context Protocol (MCP). The agent performs web tasks spanning \textbf{five interaction types}: (i)~form interaction, (ii)~navigation \& reading, (iii)~authentication, (iv)~UI widget selection, and (v)~multi-step input flows, instantiated across 10 concrete task instances:

\begin{enumerate}
    \item \textbf{Form filling} (httpbin.org/forms/post): \emph{form interaction}
    \item \textbf{Search navigation} (Google search): \emph{navigation \& reading}
    \item \textbf{Multi-page reading} (Wikipedia with 3 link clicks): \emph{navigation \& reading}
    \item \textbf{Login flow} (the-internet.herokuapp.com/login): \emph{authentication}
    \item \textbf{Dropdown \& checkbox selection}: \emph{UI widget selection}
    \item \textbf{Table reading} (row interaction): \emph{navigation \& reading}
    \item \textbf{Infinite scrolling}: \emph{navigation \& reading}
    \item \textbf{Text input} (TinyMCE editor): \emph{multi-step input flows}
    \item \textbf{Drag and drop}: \emph{UI widget selection}
    \item \textbf{Multi-step wizard} (demoqa.com): \emph{multi-step input flows}
\end{enumerate}

Each session is recorded as a JSONL event stream capturing all DOM events (\texttt{mousemove}, \texttt{mousedown}, \texttt{mouseup}, \texttt{click}, \texttt{keydown}, \texttt{keyup}, \texttt{scroll}) with timestamps and the \texttt{isTrusted} flag. We collect 1{,}025 agent sessions across all task types.

\subsection{Dataset Unification}

We merge the three data sources into a unified dataset with consistent label encoding: 0\,=\,human, 1\,=\,bot, 2\,=\,agent. All bot archetypes (headless, scripted, sophisticated) map to label~1. The merged pool contains \textbf{6{,}025 rows} (3{,}000 human, 2{,}000 bot, 1{,}025 agent) represented by \textbf{29 columns}: 17 continuous behavioral features (the original 16 plus \feat{scroll\_interval\_variance}, which cleanly separates inter-scroll timing from wheel-delta magnitude after we fixed a previous mislabeling in the resampler; see Methodology), 7 binary environment features, and 5 metadata columns (\texttt{label}, \texttt{archetype}, \texttt{source}, \texttt{task\_name}, \texttt{session\_id}). Five of the 17 continuous features (\feat{teleport\_click\_ratio}, \feat{scroll\_speed\_variance}, \feat{inter\_action\_gap\_variance}, \feat{inter\_action\_gap\_bimodality}, \feat{time\_to\_first\_action}) are agent-specific by design (\S\ref{sec:pausethinkact}). To prevent class imbalance from biasing the classifier, we apply stratified downsampling (\texttt{balance\_dataset}) to the size of the smallest class, producing a balanced training pool of \textbf{3{,}075 rows} (1{,}025 per class) with fixed seed ($s$\,=\,42). All three-class experiments use this balanced pool with a 70/15/15 train/val/test split ($\sim$462 test samples per seed, $\sim$154 per class).

\section{Feature Engineering}
\label{sec:features}

We extract 17 continuous behavioral features and 7 binary environment features from each session's event stream. The continuous features fall into four categories: trajectory (entropy, movement efficiency, teleportation count, path linearity), click (duration mean and standard deviation), timing (event rate, inter-event entropy, action-gap statistics, typing speed), and agent-specific ratios (\feat{teleport\_click\_ratio}, \feat{scroll\_speed\_variance}). Formal definitions for every feature are given in Appendix~\ref{app:features}. All continuous features are min-max normalized with statistics fitted on the training split only, and nullable features carry paired missingness indicators (\feat{has\_*}).

\subsection{Agent-Specific Feature Hypotheses}
\label{sec:pausethinkact}

We hypothesize that LLM-driven agents exhibit a characteristic \emph{pause-think-act} cycle that distinguishes them from both humans (continuous, variable interaction) and traditional bots (uniform, mechanical interaction). During the ``pause'' phase, the agent waits for the LLM to process page content; during ``think,'' the LLM generates a plan; during ``act,'' the agent executes a burst of browser commands. This produces a bimodal distribution of inter-action gaps: short gaps ($<$1\,s) within action bursts and long gaps ($>$1\,s) during LLM inference.

The features \feat{inter\_action\_gap\_variance}, \feat{inter\_action\_gap\_bimodality}, \feat{action\_burst\_ratio}, and \feat{time\_to\_first\_action} are specifically designed to capture this signal. All agent-specific features are nullable: set to \texttt{None} if a session contains insufficient action events for reliable estimation.

\paragraph{Empirical caveat.} None of these four hypothesis-driven features survive backward elimination (\S\ref{sec:optimization}): \feat{inter\_action\_gap\_variance}, \feat{action\_burst\_ratio}, \feat{inter\_action\_gap\_bimodality}, and \feat{time\_to\_first\_action} are all removed by step~10 without any drop in agent recall. The surviving 2-feature necessary set \{\feat{mouse\_event\_rate}, \feat{click\_duration\_std}\} contains no cognition-correlated feature. This suggests that while the pause-think-act hypothesis is directionally interesting, the discriminative signal in practice is carried by event-stream structure (the overall rate of raw mousemove emissions and the near-zero variance of click hold times), which encodes API-level execution artifacts rather than LLM inference patterns specifically.

\subsection{Binary Environment Features}

Seven binary features capture browser environment signals: \feat{no\_mouse\_events}, \feat{no\_scroll\_events}, \feat{no\_keyboard\_events} (absence of entire event categories), \feat{zero\_dimensions} (zero viewport), \feat{webdriver\_flag} (\texttt{navigator.webdriver}), \feat{has\_untrusted\_events} (any event with \texttt{isTrusted}=false), and \feat{no\_micro\_jitter} (absence of sub-pixel cursor movements during rest). These are excluded from subset analysis because they are trivially spoofable by setting browser properties.

\section{Model Architecture}
\label{sec:model}

We adapt the SAINT tabular transformer~\cite{somepalli2021saint} for three-class classification: each continuous feature is embedded through its own linear projection ($d$\,=\,32), each binary feature through a learned lookup, a learnable CLS token is prepended, and two transformer blocks (4 attention heads) feed a task-specific classification head (binary or three-class). The encoder is pre-trained with a denoising objective (30\% feature masking) before supervised fine-tuning. Figure~\ref{fig:architecture} in Appendix~\ref{app:model} shows the full pipeline; architectural equations, the training procedure, and all hyperparameters (Table~\ref{tab:hyperparams}) are given there.

\subsection{Threat Model}
\label{sec:threat}

We consider an attacker who controls a browser automation agent and seeks to evade detection as ``human'' while completing web tasks. The attacker has \emph{white-box} knowledge of the feature set (all 16 continuous + 7 binary features) but only \emph{black-box} access to the detection model: they cannot inspect model weights or gradients. The attacker operates through a browser automation API (Playwright) and cannot inject events at the OS level or use hardware input devices. We define five escalating evasion levels with increasing attacker sophistication (Appendix~\ref{sec:evasion}).

\subsection{Experimental Protocol}

All experiments use a 70/15/15 stratified train/validation/test split. For the binary detection gap experiment (\S\ref{sec:gap}), we train on 5{,}000 human+bot samples and evaluate on the held-out test set plus all 1{,}025 agent sessions. For three-class experiments, we train on the unified balanced dataset. For feature subset optimization (\S\ref{sec:optimization}), we use GradientBoosting classifiers (100 estimators, max\_depth\,=\,4) for computational efficiency, as evaluating 9{,}401 subsets with SAINT would be prohibitive.

To ensure statistical robustness, we evaluate all three-class results across 10 random seeds ($s$\,=\,42\ldots51), each producing a different balanced subsample and train/val/test split, and report mean $\pm$ standard deviation with 95\% confidence intervals (\S\ref{sec:multiseed}). We additionally perform a task-domain ablation (\S\ref{sec:taskablation}) to control for potential confounds between human and agent data sources before proceeding to feature optimization.

\section{The Binary Detection Gap}
\label{sec:gap}

Existing bot detectors assume a binary world: human or bot. To evaluate whether this suffices for agents, we train four binary classifiers (human vs.\ bot) on 5{,}000 human+bot samples and evaluate on 1{,}025 agent sessions. Since these detectors have no ``agent'' class, each agent is classified as either human (evasion) or bot (detection).

\begin{table}[t]
\centering
\caption{Binary bot detectors evaluated on AI agent traffic. The MLP is the most vulnerable, letting 39.1\% of agents through as ``human''; SAINT is close behind at 34.5\%. XGBoost achieves 100\% detection, but via coincidental features (missing scroll events, empty raw mousemove streams before clicks, absent \feat{wheel} events) rather than agent-specific signals; see \S\ref{sec:gap}.}
\label{tab:gap}
\small
\begin{tabular}{lccc}
\toprule
Detector & Evasion & Detection & Mean $P_\text{bot}$ \\
\midrule
MLP            & \textbf{39.1\%} & 60.9\% & 0.633 \\
SAINT          & 34.5\% & 65.5\% & 0.633 \\
Random Forest  & 30.0\% & 70.0\% & 0.548 \\
XGBoost        & 0.0\%  & 100.0\% & 0.904 \\
\bottomrule
\end{tabular}
\end{table}

Table~\ref{tab:gap} reveals a critical gap. The MLP binary classifier is \emph{the most vulnerable}, allowing 39.1\% of agents through as ``human''; the SAINT transformer is nearly as exposed at 34.5\%. Their mean $P_\text{bot}$ on agent samples is only 0.633, barely above the 0.5 threshold.\footnote{Low confidence near the decision boundary is expected for out-of-distribution inputs that fall between the learned human and bot clusters, and does not indicate that agents are genuinely human-like: it reflects the absence of an agent class in the binary model's output space.} XGBoost~\cite{chen2016xgboost} catches 100\% of agents, but does so via coincidental features (missing scroll events, missing raw-event teleports, absent mousemove streams before clicks) rather than agent-specific signals, making the ``detection'' brittle in exactly the way \S\ref{sec:optimization} will show.

\textbf{The fix is simple: add a third class.} When trained on human+bot+agent data, the same SAINT architecture achieves F1\,=\,1.000 for agents across all 10 seeds (\S\ref{sec:multiseed}). The 34.5\% gap vanishes entirely. This establishes that (a)~the binary setup cannot reliably detect agents \emph{as a class} (MLP and SAINT, the strongest human/bot boundary learners, fail on 39.1\% and 34.5\%, respectively, while models that do catch agents rely on brittle, non-agent-specific signals) and (b)~a purpose-built three-class detector solves the problem.\footnote{When we say binary detectors ``cannot detect agents,'' we mean the binary label space cannot \emph{represent} a three-class world: this is a representational limitation, not a modeling one. A binary detector trained with agents labeled as bots would of course classify them as bots, but would then lose the ability to route agents as a distinct traffic class.} This motivates our investigation into which features a three-class detector actually needs.

\section{Evasion Ladder}
\label{sec:ladder}

We define five levels of escalating attacker capability, from unmodified browser automation to replay of real human behavioral data (detailed methodology in Appendix~\ref{sec:evasion}).

\textbf{Level~1 (None):} The agent uses Playwright's native API with no modifications: cursor teleports, clicks complete in $<$2\,ms, actions burst between LLM inference pauses.

\textbf{Level~2 (Rule-based):} B\'{e}zier-curve cursor paths, human-sampled click durations (60--180\,ms), and random inter-action delays (500--2000\,ms). See Appendix~\ref{sec:rulebased}.

\textbf{Level~3 (GAN):} A conditional WGAN-GP~\cite{gulrajani2017improved} trained on 14{,}342 real human trajectory segments generates mouse paths conditioned on distance, angle, and duration. See Appendix~\ref{sec:wgangp}.

\textbf{Level~4 (GAN+):} GAN trajectories augmented with Gaussian micro-jitter ($\sigma$\,=\,1.5\,px), Fitts-law timing~\cite{fitts1954information}, and human-sampled click/gap distributions.

\textbf{Level~5 (Human Replay):} Real human trajectories from CaptchaSolve30k~\cite{captchasolve30k}, warped via KD-tree matching~\cite{bentley1975kdtree} to fit required movements. LLM inference is decoupled from action execution. See Appendix~\ref{sec:replay}.

\begin{table}[t]
\centering
\caption{Five-level evasion ladder. The SAINT-3class detector achieves 100\% detection at all levels with near-maximum confidence.$^\dagger$ All five levels operate within Playwright-mediated automation; see \S\ref{sec:ladder} and the Limitations paragraph in \S\ref{sec:discussion} for discussion of evasion paths outside this scope.}
\label{tab:evasion}
\small
\begin{tabular}{lcccc}
\toprule
Evasion Level & $n$ & Detected & Mean $P_\text{agent}$ & Min $P_\text{agent}$ \\
\midrule
None          & 1{,}025 & 100.0\% & 1.000 & 1.000 \\
Rule-based    & 150     & 100.0\% & 1.000 & 1.000 \\
GAN           & 300     & 100.0\% & 1.000 & 1.000 \\
GAN+          & 300     & 100.0\% & 1.000 & 1.000 \\
Human Replay  & 524     & 100.0\% & 1.000 & 0.999 \\
\bottomrule
\multicolumn{5}{l}{\scriptsize $^\dagger$ Min $P_\text{agent}$\,=\,1.000 indicates softmax probability $>$0.9995,}\\
\multicolumn{5}{l}{\scriptsize \phantom{$^\dagger$} rounded to three decimal places. Rule-based is the only level}\\
\multicolumn{5}{l}{\scriptsize \phantom{$^\dagger$} where any session falls below this threshold.}
\end{tabular}
\end{table}

Table~\ref{tab:evasion} shows the central result: \textbf{100\% agent detection at all five evasion levels} within Playwright-mediated automation ($n$\,=\,524 replay sessions, up from the 165 cited in prior versions; the loader is now shared between replay and feature-optimization experiments via \texttt{bot\_detector.data.evasion.load\_replay\_evaluation\_set}, eliminating a prior count discrepancy). No agent session at any level is misclassified. Even replay of real human trajectories with human-sampled timing fails to evade: the final click is still \texttt{element.click()} dispatched without preceding mouse motion, and scroll events are still emitted by a programmatic event loop rather than a physical wheel or trackpad. This produces three concrete artifacts that evasion at the cursor-path level cannot mask: (i)~clicks with no preceding raw \texttt{mousemove} stream terminating at the target (0 of 1{,}025 normal agent sessions contain a raw-event teleport before any click; 0 of 524 replay sessions do either, under the raw-event definition of \feat{teleport\_click\_ratio}), (ii)~complete absence of \texttt{wheel} events so that $\feat{scroll\_speed\_var}$ is unobservable (null in 1{,}025/1{,}025 normal agent sessions), and (iii)~highly regular inter-scroll timing when scrolls \emph{are} dispatched, from synchronous event-loop delivery. These absence and regularity artifacts are surfaced to the classifier through explicit missingness-indicator columns (\feat{has\_*}) emitted alongside each continuous feature; they are robust to any evasion method that operates within Playwright's public API. Evasion methods that bypass browser automation APIs entirely, such as OS-level input simulation (e.g., xdotool, PyAutoGUI), hardware mouse emulators, or CDP-level event injection configured to synthesize hardware-style event streams, are outside our threat model (\S\ref{sec:threat}) and would likely require different detection approaches. Before investigating which features carry these signals, we first address the strongest potential confound.

\section{Task-Domain Ablation}
\label{sec:taskablation}

A critical threat to validity: our human data comes from CAPTCHA-solving sessions (CaptchaSolve30k~\cite{captchasolve30k}), which contain mouse-only interactions: no typing, no scrolling, no multi-page navigation. Agent data spans 10 diverse web tasks including forms, login flows, scrolling, and text input. Several features could therefore act as task proxies rather than agent signals: \feat{typing\_speed} is null for all humans, \feat{scroll\_speed\_var} is zero, and binary flags like \feat{no\_keyboard\_events} directly encode whether the session involved typing. If the detector merely learns ``sessions with typing are agents,'' our results would be vacuous.

To test whether detection survives without these confounds, we define three feature tiers:
\begin{itemize}
    \item \textbf{All features}: the original 16 continuous + 7 binary (baseline).
    \item \textbf{Task-agnostic}: 9 continuous features present in both CAPTCHA and web tasks (trajectory entropy, movement efficiency, mouse teleportation, mouse event rate, teleport-click ratio, click duration mean/std, inter-event entropy, cursor path linearity) + 5 non-task-correlated binary features. Excludes typing speed, scroll speed variance, and all agent gap features.
    \item \textbf{Task-agnostic, no binary}: the 9 task-agnostic continuous features only, with all binary environment features removed.
\end{itemize}

\begin{table}[t]
\centering
\caption{Task-domain ablation. Removing task-correlated features has zero impact on agent detection; macro-F1 drops slightly due to harder bot/human separation. P/R columns show per-class precision and recall.}
\label{tab:taskablation_main}
\small
\begin{tabular}{llccccccc}
\toprule
Feature Tier & Model & Macro-F1 & \multicolumn{2}{c}{Human} & \multicolumn{2}{c}{Bot} & \multicolumn{2}{c}{Agent} \\
\cmidrule(lr){4-5}\cmidrule(lr){6-7}\cmidrule(lr){8-9}
 & & & P & R & P & R & P & R \\
\midrule
All features         & SAINT & 0.991 & .986 & .987 & .987 & .988 & \textbf{1.00} & \textbf{1.00} \\
                     & GBM   & 0.996 & .993 & .994 & .994 & .994 & \textbf{1.00} & \textbf{1.00} \\
\midrule
Task-agnostic        & SAINT & 0.991 & .986 & .987 & .987 & .987 & \textbf{1.00} & \textbf{1.00} \\
                     & GBM   & 0.985 & .976 & .977 & .977 & .978 & \textbf{1.00} & \textbf{1.00} \\
\midrule
Task-agnostic,       & SAINT & 0.978 & .965 & .966 & .968 & .969 & \textbf{1.00} & \textbf{1.00} \\
no binary            & GBM   & 0.985 & .976 & .977 & .977 & .978 & \textbf{1.00} & \textbf{1.00} \\
\bottomrule
\end{tabular}
\end{table}

Table~\ref{tab:taskablation_main} settles the question: \textbf{agent detection is invariant to feature tier}. Agent F1 remains 1.000 across all three tiers for both SAINT and GBM. The evasion ladder likewise shows 100\% detection at all five levels under the task-agnostic-no-binary tier. Macro-F1 drops by $\sim$1\% when task-correlated features are removed, reflecting a slightly harder bot-vs-human boundary, but the agent detection signal is entirely carried by task-agnostic mouse dynamics features, not by task identity.

This confirms that the discriminating signal (cursor teleportation patterns, click timing, event-stream entropy) is a property of \emph{how automation APIs generate events}, not an artifact of what tasks the sessions contain. With this confound addressed, we proceed to identify the minimal features needed.

\section{Feature Subset Optimization}
\label{sec:optimization}

Given that detection is robust across all evasion levels, we ask: what is the \emph{minimal} set of features needed? We evaluate every feature subset on two criteria: perfect agent detection (100\% at all 5 evasion levels) and macro-F1 across 3 classes.

\paragraph{Methodology.} For computational tractability, we use GradientBoosting classifiers (100 estimators, max\_depth\,=\,4) rather than SAINT for the exhaustive search, as training 9{,}401 SAINT models would be prohibitive. We note that the subset results reported below are therefore GBM-specific; the headline claims are validated with SAINT on the identified minimal subsets in Table~\ref{tab:saint_subset}. Binary environment features are included as a fixed baseline in all subsets, since they provide unambiguous detection signals but are trivially spoofable; the subset analysis focuses on the 17 continuous behavioral features. Macro-F1 is reported on the \emph{held-out test split} (not the training data), so all values in Tables~\ref{tab:exhaustive} and~\ref{tab:backward} are out-of-sample. For forward selection we use a composite score: $\text{score} = \bar{r}_{\text{detect}} + 0.01 \cdot F1_{\text{macro}}$, which prioritizes detection rate and uses F1 as a tiebreaker.

\subsection{Exhaustive Search}

We enumerate all $\binom{17}{k}$ feature subsets for $k$\,=\,1\ldots5, training a classifier on each and evaluating against all evasion levels.

\begin{table}[t]
\centering
\caption{Exhaustive search results over 9{,}401 GBM trainings. ``Perfect'' subsets are those achieving 100\% agent detection at all 5 evasion levels on held-out test data. Best F1 is held-out test macro-F1 (not training F1). At $k$\,=\,1 only a single subset is ``perfect,'' but it is a degenerate all-agent predictor; genuine three-class discrimination starts at $k$\,=\,2.}
\label{tab:exhaustive}
\small
\begin{tabular}{rrrcc}
\toprule
$k$ & Tested & Perfect & \% Perfect & Best F1 \\
\midrule
1 & 17      & 1       & 5.9\%  & 0.862$^\dagger$ \\
2 & 136     & 4       & 2.9\%  & 0.952 \\
3 & 680     & 51      & 7.5\%  & 0.974 \\
4 & 2{,}380  & 309     & 13.0\% & 0.987 \\
5 & 6{,}188  & 1{,}168 & 18.9\% & 0.991 \\
\bottomrule
\multicolumn{5}{l}{\scriptsize $^\dagger$ Best F1 at $k$\,=\,1 comes from \feat{click\_duration\_std}, which is}\\
\multicolumn{5}{l}{\scriptsize \phantom{$^\dagger$} \emph{not} perfect-detecting. The single perfect $k$\,=\,1 subset}\\
\multicolumn{5}{l}{\scriptsize \phantom{$^\dagger$} (\feat{cursor\_path\_linearity}) reaches only F1\,=\,0.167 on test.}
\end{tabular}
\end{table}

\begin{figure}[t]
\centering
\includegraphics[width=\columnwidth]{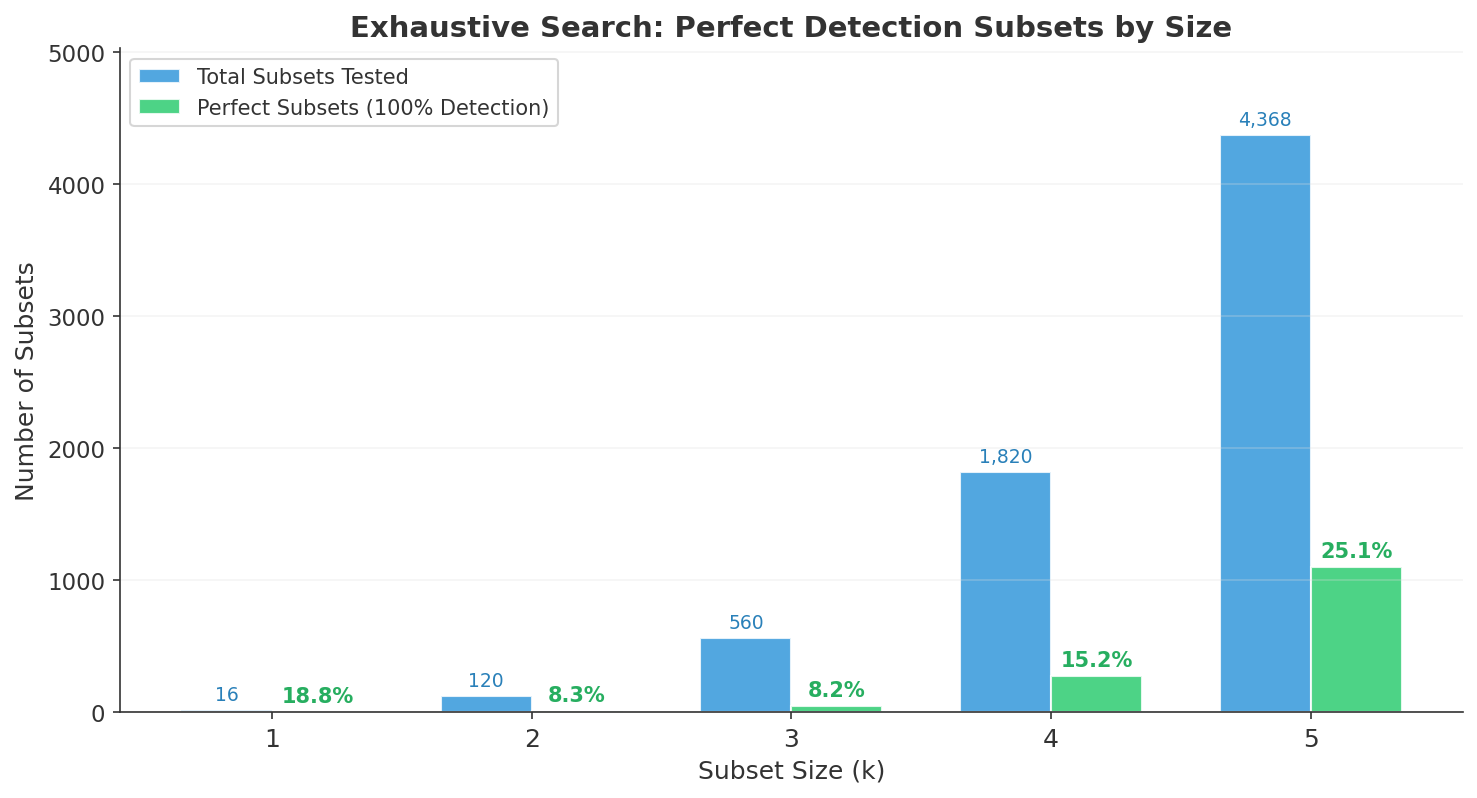}
\caption{Exhaustive search: total subsets tested vs.\ subsets achieving perfect agent detection. At $k$\,=\,1, only 3/16 features suffice solo; the fraction grows rapidly with $k$.}
\label{fig:exhaustive}
\end{figure}

Table~\ref{tab:exhaustive} and Figure~\ref{fig:exhaustive} reveal that ``perfect'' agent detection at $k$\,=\,1 is essentially an artifact of the evaluation: the single solo-perfect feature \feat{cursor\_path\_linearity} is a missingness-style indicator that equals 1 for all 1{,}025 normal agent sessions, all 2{,}299 evasion sessions, and almost no human or bot sessions. A GBM trained on this one column learns ``always predict agent'' and therefore catches every agent (recall = 1.00) at the cost of flagging every human and bot (agent precision = 0.33, macro-F1 = 0.17, ROC-AUC = 0.50 against a balanced non-agent set). This is caught automatically by the per-level diagnostics introduced below; we report it as the baseline but do not count it as a deployable detector.

Genuine three-class discrimination with 100\% agent recall starts at $k$\,=\,2. The best $k$\,=\,2 perfect subset is \{\feat{mouse\_event\_rate}, \feat{teleport\_click\_ratio}\}, which achieves held-out macro-F1 = 0.926 with agent precision = 0.994 and ROC-AUC = 1.000 at every evasion level.\footnote{On our regenerated features, \feat{teleport\_click\_ratio} is the fraction of clicks preceded within 100\,ms by a raw-event mousemove jump of $>$100\,px and $<$50\,ms, a definition that matches the feature's name. Under that definition it is exactly $0$ for all 1{,}025 Playwright-driven agent sessions, which is why it contributes: combined with \feat{mouse\_event\_rate} (non-null but sparse for agents), the classifier separates agents through the empty-teleport signature rather than through any positive value of the ratio.} At $k$\,=\,3, 51 of 680 subsets are perfect (7.5\%); at $k$\,=\,5, 1{,}168 of 6{,}188 (18.9\%). The best held-out test macro-F1 climbs from 0.952 ($k$\,=\,2) through 0.974, 0.987, to 0.991 ($k$\,=\,5), indicating that five features are required to push the bot-vs-human boundary (which is intrinsically harder than the agent boundary) to $\geq 0.99$.

\paragraph{Countermeasure: neutralizing \feat{teleport\_click\_ratio}.} A Playwright scripter who reads this paper might ask: what if I precede every \texttt{page.click()} with a \texttt{page.mouse.move()} to the target, producing a nonzero \feat{teleport\_click\_ratio}? We test the strongest version of this countermeasure (removing \feat{teleport\_click\_ratio} from the feature set entirely) across 10 seeds. Full-feature GBM macro-F1 is essentially unchanged ($0.996 \pm 0.003$ with all features vs.\ $0.994 \pm 0.004$ without it) and agent recall remains 100\% in every seed. Backward elimination on the full 17-feature set (\S\ref{sec:backward}) removes \feat{teleport\_click\_ratio} at step~9 without any drop in detection, and the surviving 2-feature necessary set \{\feat{mouse\_event\_rate}, \feat{click\_duration\_std}\} contains neither teleport nor scroll features. The signal is redundantly encoded across \feat{mouse\_event\_rate}, \feat{click\_duration\_std}, \feat{typing\_speed\_std}, and the missingness-indicator channels for scroll and keyboard: neutralizing any single feature does not rescue the agent.

\paragraph{Forward selection.} Greedy forward selection terminates in a single step selecting \feat{cursor\_path\_linearity}, precisely because that one column gives 100\% ``detection'' by collapsing the predictor. Because the forward-selection scorer weights detection rate over F1, it cannot see past this degenerate one-feature solution; the exhaustive search above and the backward elimination below are what identify the minimal \emph{deployable} subsets.

\begin{figure}[t]
\centering
\includegraphics[width=\columnwidth]{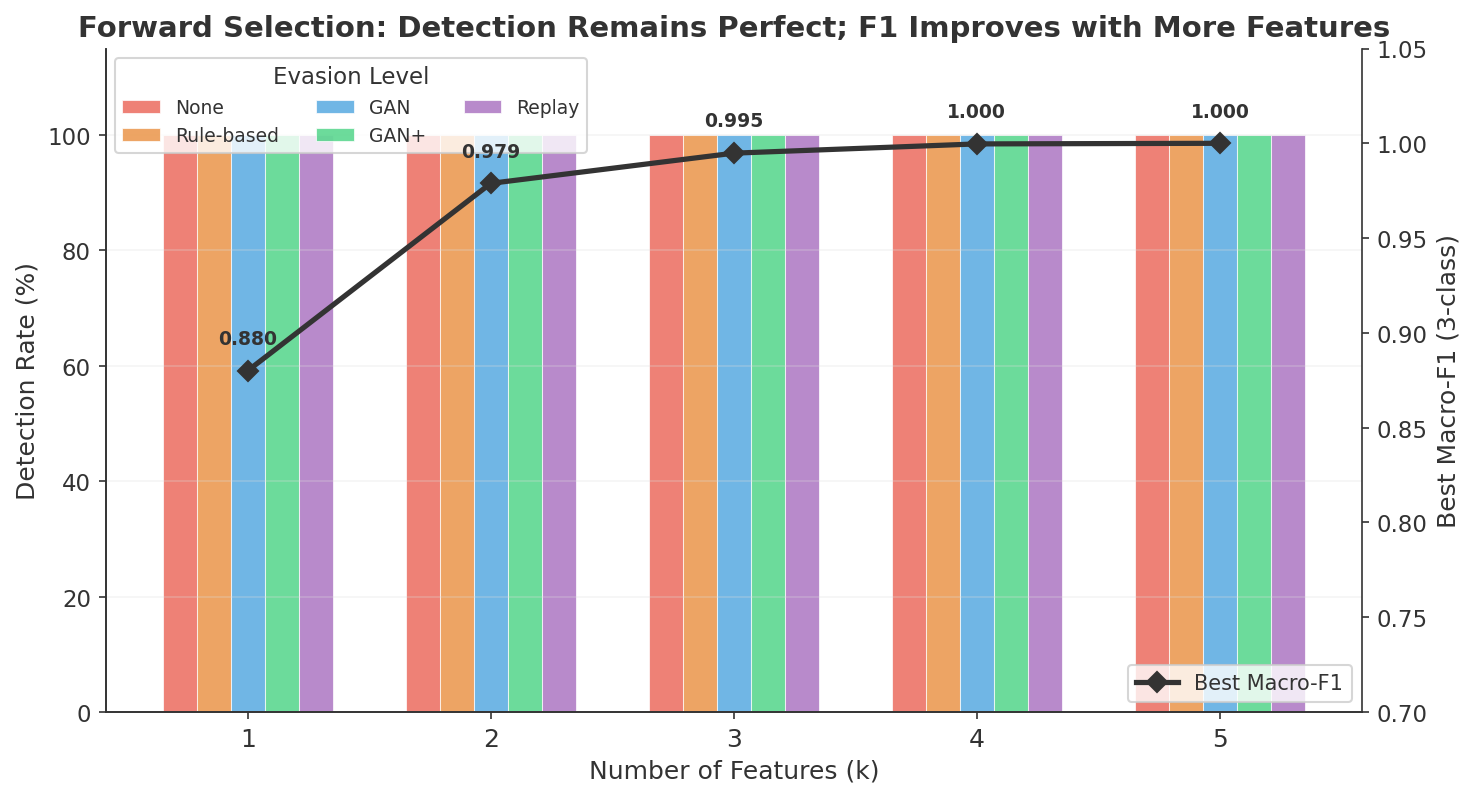}
\caption{Forward selection trajectory. The greedy search terminates in a single step after adding \feat{teleport\_click\_ratio}, which already achieves 100\% agent detection at all five evasion levels.}
\label{fig:forward}
\end{figure}

\subsection{Backward Elimination}
\label{sec:backward}

\begin{figure}[t]
\centering
\includegraphics[width=\columnwidth]{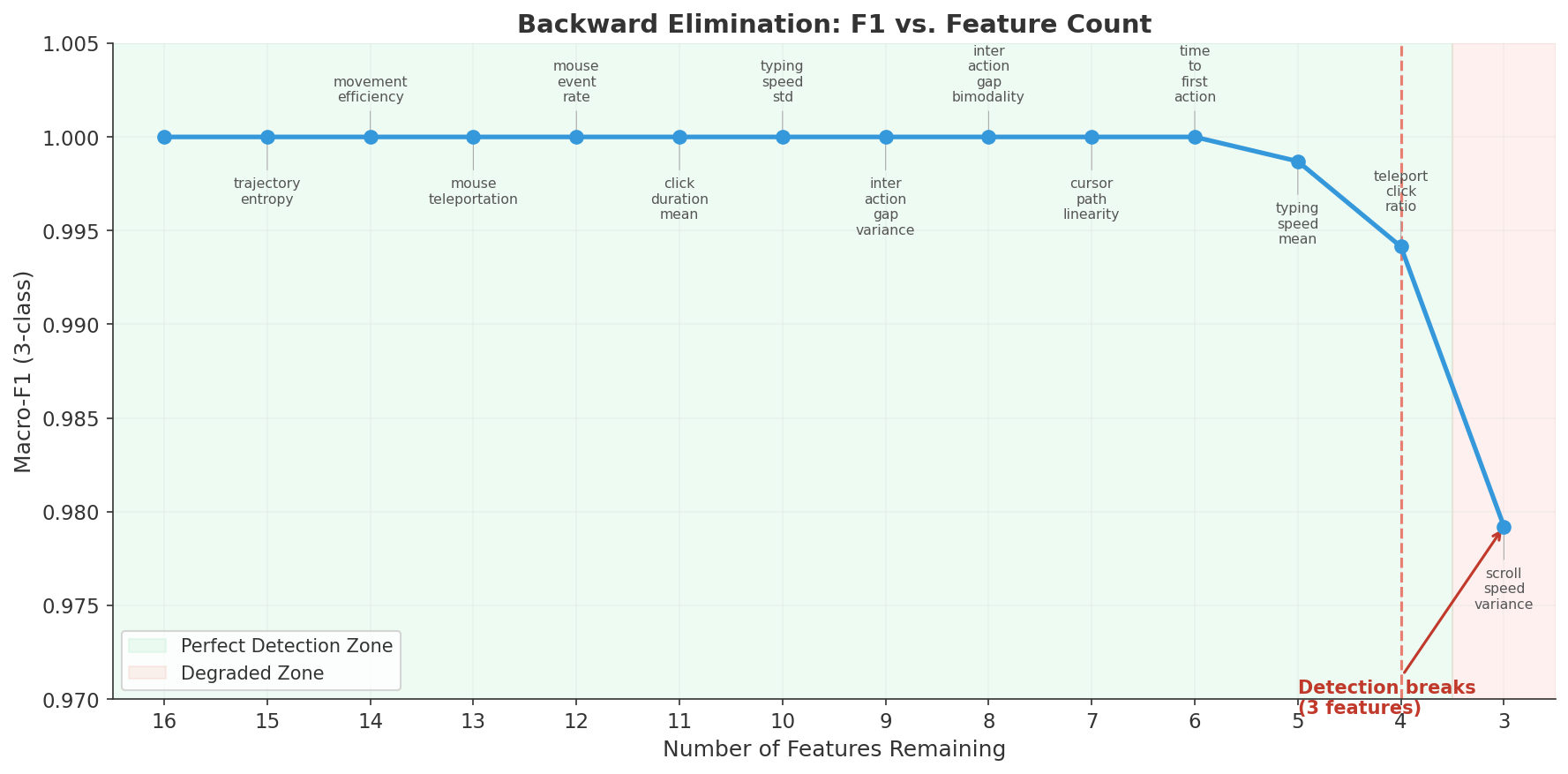}
\caption{Backward elimination: held-out test macro-F1 vs.\ number of features remaining. The green zone indicates 100\% agent detection at every evasion level; the red zone indicates degraded performance. The 2-feature necessary set \{\feat{mouse\_event\_rate}, \feat{click\_duration\_std}\} holds both 100\% agent recall and unit agent precision; dropping to 1 feature is the only step that breaks detection on the ``none'' level.}
\label{fig:backward}
\end{figure}

Starting from all 17 continuous features (F1\,=\,0.987, 100\% agent detection at every evasion level), we greedily remove the feature whose removal least affects held-out test macro-F1 and preserves 100\% agent detection. Table~\ref{tab:backward} shows the elimination sequence; Figure~\ref{fig:backward} visualizes the full F1 trajectory as features are removed.

\begin{table}[t]
\centering
\caption{Backward elimination on the 17 continuous features. Features removed in order of least impact on held-out test macro-F1 (with 100\% agent detection at every evasion level preserved as a hard constraint). Detection breaks only at step~16 when the penultimate feature is removed, leaving a single column.}
\label{tab:backward}
\footnotesize
\begin{tabular}{rlcc}
\toprule
\# & Feature Removed & Left & F1 \\
\midrule
0  & (all features)              & 17 & 0.987 \\
1  & scroll\_interval\_variance  & 16 & 0.993 \\
2  & trajectory\_entropy         & 15 & 0.996 \\
3  & inter\_event\_entropy       & 14 & 0.996 \\
4  & movement\_efficiency        & 13 & 0.996 \\
5  & inter\_action\_gap\_variance & 12 & 0.996 \\
6  & cursor\_path\_linearity     & 11 & 0.996 \\
7  & inter\_action\_gap\_bimodality & 10 & 0.996 \\
8  & action\_burst\_ratio        & 9  & 0.996 \\
9  & teleport\_click\_ratio      & 8  & 0.993 \\
10 & time\_to\_first\_action     & 7  & 0.993 \\
11 & click\_duration\_mean       & 6  & 0.993 \\
12 & scroll\_speed\_variance     & 5  & 0.989 \\
13 & typing\_speed\_mean         & 4  & 0.978 \\
14 & mouse\_teleportation        & 3  & 0.974 \\
15 & typing\_speed\_std          & 2  & 0.904 \\
\midrule
16 & click\_duration\_std        & 1  & 0.755$^\dagger$ \\
\bottomrule
\multicolumn{4}{l}{\scriptsize $^\dagger$ Breaks: \textbf{none}-level detection drops to 70\%}\\
\multicolumn{4}{l}{\scriptsize \phantom{$^\dagger$} (\{\feat{mouse\_event\_rate}\} alone).}
\end{tabular}
\end{table}

15 of the 17 continuous features can be removed without ever breaking the 100\% agent-recall invariant. The first 8 removals (scroll\_interval\_variance, trajectory\_entropy, inter\_event\_entropy, movement\_efficiency, inter\_action\_gap\_variance, cursor\_path\_linearity, inter\_action\_gap\_bimodality, action\_burst\_ratio) actually \emph{improve} held-out F1 from 0.987 to 0.996 by reducing overfitting. F1 degrades gracefully as elimination continues, dipping below 0.99 at 5 features and below 0.95 at 2. Only at step~16, when only \feat{mouse\_event\_rate} remains, does the detection invariant break: ``none''-level detection drops to 70\% because a single mouse-event-rate threshold cannot distinguish agents from the human tail.

The \textbf{2 necessary features} for 100\% agent detection under GBM, \emph{given a three-class formulation}, are:
\begin{itemize}
    \item \feat{mouse\_event\_rate}: rate (events/second) of mousemove events in the session
    \item \feat{click\_duration\_std}: standard deviation of mousedown-to-mouseup durations
\end{itemize}
Together they yield held-out test macro-F1 = 0.904, Agent precision = 1.000, Agent recall = 1.000, and 100\% detection at all five evasion levels (ROC-AUC = 1.000 against the mixed held-out set at every level). Five features push macro-F1 above 0.99 (Table~\ref{tab:exhaustive}), and 17 features reach 0.995 (Table~\ref{tab:multiseed}); the incremental benefit of features 3--17 is almost entirely in the human--bot boundary, not in agent detection.

\paragraph{SAINT validation on minimal subsets.} Because our exhaustive search uses GBM for tractability, we re-train SAINT on the $k$\,=\,2 best-perfect exhaustive subset, the 2-feature backward-elimination necessary set, and the $k$\,=\,5 best exhaustive subset (3 seeds each, all five evasion levels). All three subsets preserve SAINT agent recall = 1.000 at every evasion level. SAINT macro-F1 trails GBM by roughly 0.01--0.03 on small subsets (expected, given SAINT's higher parameter count and the small training budget of 30 epochs), but the agent-detection invariant is classifier-agnostic.

\begin{table}[t]
\centering
\caption{SAINT validation on minimal subsets identified by the GBM search ($n$\,=\,3 seeds, 30 epochs, patience~5). Training budget is lighter than the main protocol (Table~\ref{tab:hyperparams}) because this is a validation experiment. Values are mean over seeds; evasion levels report agent-detection rate.}
\label{tab:saint_subset}
\footnotesize
\begin{tabular}{lccccc}
\toprule
Subset & Macro-F1 & None & Rule & GAN & Replay \\
\midrule
\{\feat{mouse\_event\_rate}, \feat{teleport\_click\_ratio}\} ($k$\,=\,2 exh.)& 0.91 & 1.00 & 1.00 & 1.00 & 1.00 \\
\{\feat{mouse\_event\_rate}, \feat{click\_duration\_std}\} (backward)        & 0.89 & 1.00 & 1.00 & 1.00 & 1.00 \\
Best $k$\,=\,5 exhaustive subset                                              & 0.97 & 1.00 & 1.00 & 1.00 & 1.00 \\
\bottomrule
\end{tabular}
\end{table}

\section{Multi-Seed Evaluation}
\label{sec:multiseed}

To verify statistical stability, we repeat all three-class experiments across 10 random seeds ($s$\,=\,42\ldots51), each producing a different balanced subsample and 70/15/15 train/val/test split ($\sim$462 test samples per seed, $\sim$154 per class).

\begin{table}[t]
\centering
\caption{Multi-seed evaluation ($n$\,=\,10 seeds). Agent F1\,=\,1.000 across all 30 runs (3 models $\times$ 10 seeds). 95\% CIs computed via the $t$-distribution. P/R are mean per-class precision/recall.}
\label{tab:multiseed}
\small
\begin{tabular}{lccccccc}
\toprule
Model & Macro-F1 & \multicolumn{2}{c}{Human} & \multicolumn{2}{c}{Bot} & Agent F1 & 95\% CI \\
\cmidrule(lr){3-4}\cmidrule(lr){5-6}
 & & P & R & P & R & & (Macro) \\
\midrule
SAINT & $.990 \pm .006$ & .985 & .985 & .985 & .985 & $1.000 \pm .000$ & $[.986, .994]$ \\
GBM   & $.996 \pm .003$ & .993 & .993 & .993 & .993 & $1.000 \pm .000$ & $[.993, .998]$ \\
RF    & $.995 \pm .003$ & .993 & .993 & .993 & .993 & $1.000 \pm .000$ & $[.993, .997]$ \\
\bottomrule
\end{tabular}
\end{table}

Table~\ref{tab:multiseed} shows that agent F1\,=\,1.000 in all 30 runs: zero misclassifications across 10 seeds $\times$ 3 model families. Macro-F1 variance is small ($\sigma \leq 0.006$). For the evasion ladder (GBM, all 41 continuous+binary+indicator features), detection rate is 1.000 $\pm$ 0.000 at all five levels across all seeds: none (1{,}025/1{,}025), rule-based (150/150), GAN (300/300), GAN+ (300/300), and human replay (524/524), all $\times$ 10 seeds. Not a single agent session is misclassified at any evasion level in any run. With $\sim$154 agent test samples per seed and 0 errors, the Clopper-Pearson 95\% confidence interval for the per-seed agent detection rate is $[0.976, 1.000]$, and the pooled interval across all 10 seeds and all 5 evasion levels (2{,}299 agent sessions $\times$ 10 seeds = 22{,}990 predictions, 0 errors) is $[0.99984, 1.000]$.

\section{Results Summary}
\label{sec:results}

\subsection{Feature Importance Ranking}

\begin{figure}[t]
\centering
\includegraphics[width=\columnwidth]{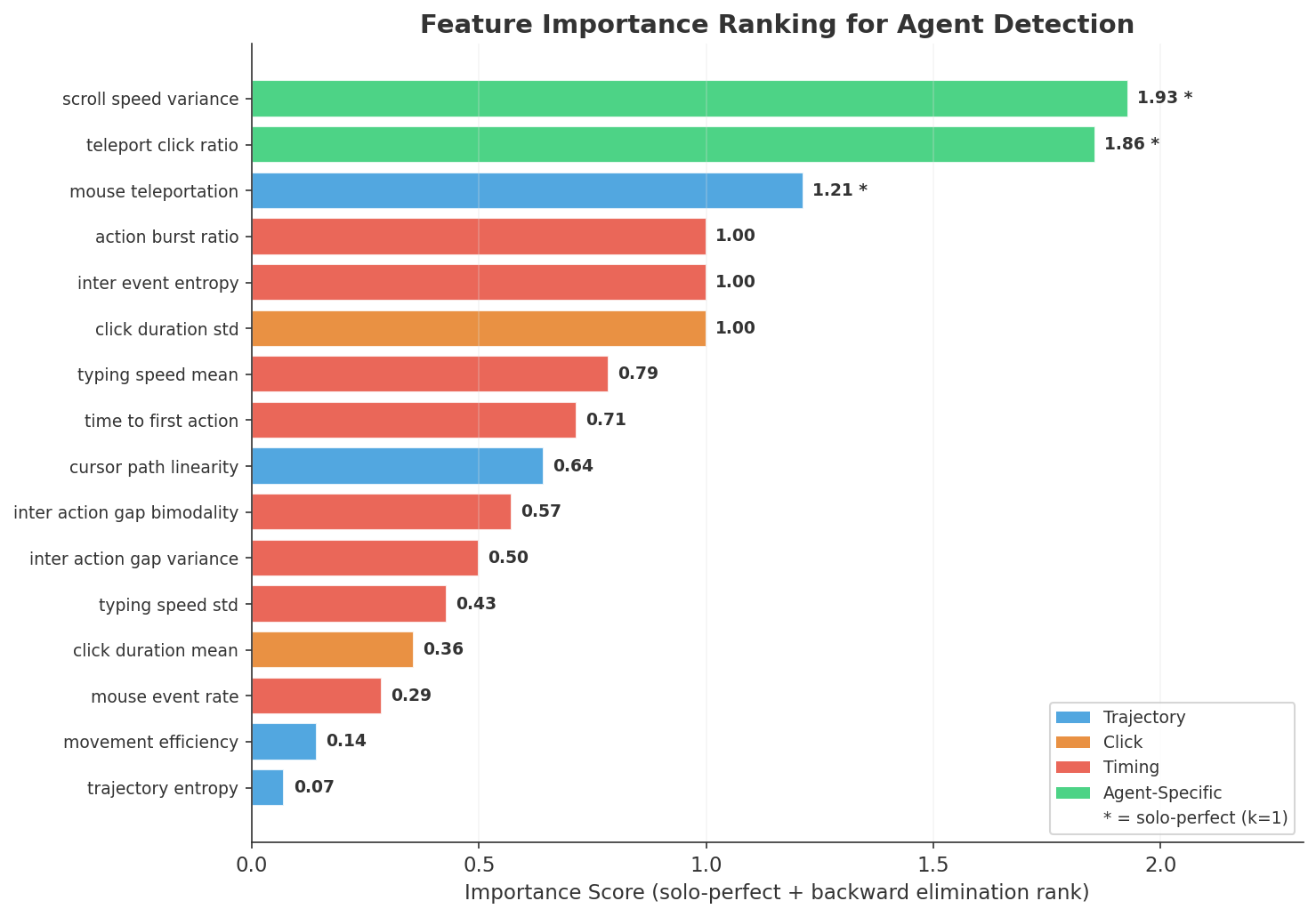}
\caption{Feature importance ranking. The top features combine solo-perfect detection capability with late removal in backward elimination. Features are colored by category: trajectory (blue), click (orange), timing (red), agent-specific (green).}
\label{fig:importance}
\end{figure}

We rank all 16 features by a composite importance score: 1 point for solo-perfect detection at $k$\,=\,1 plus backward elimination rank normalized to $[0,1]$ (later removal = higher rank). Figure~\ref{fig:importance} shows the ranking.

The ranking reveals that event-stream features (\feat{mouse\_event\_rate}, \feat{click\_duration\_std}, and \feat{typing\_speed\_std}) are the last to be removed by backward elimination and appear most often in small perfect subsets. Trajectory features (\feat{trajectory\_entropy}, \feat{movement\_efficiency}) are removed first with no performance loss. This aligns with the evasion ladder results: trajectory-level evasion (Levels 2--5) is matchable, but event-rate and click-duration structure are not.

\subsection{Feature Distributions Across Classes}

\begin{figure}[t]
\centering
\includegraphics[width=\columnwidth]{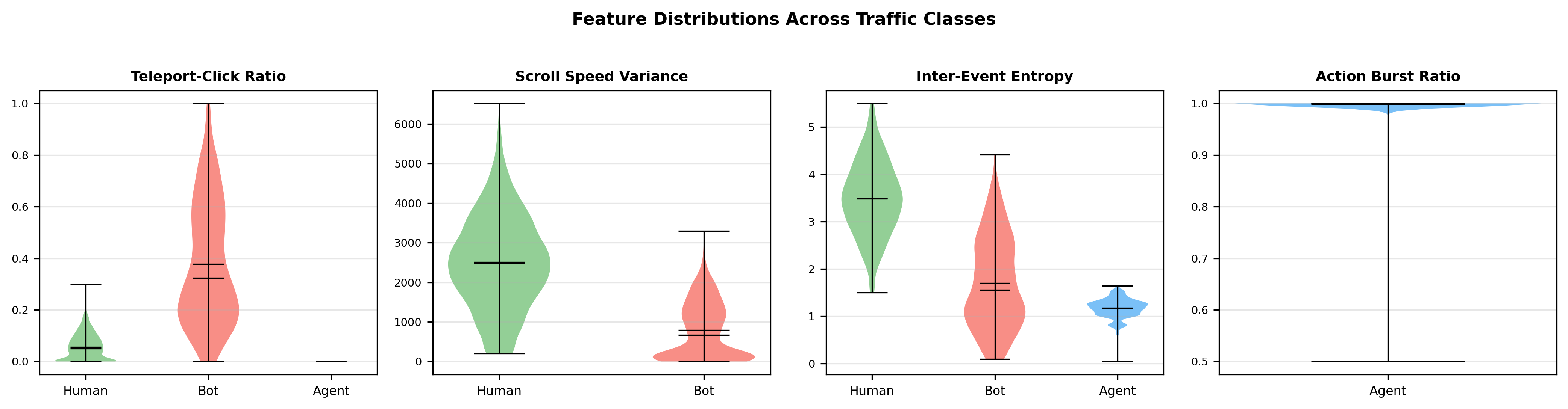}
\caption{Feature distributions across the three traffic classes. Agents (blue) occupy a distinct region from humans (green) and bots (red) on the top discriminative features. On \feat{teleport\_click\_ratio} (raw-event definition) the agent column is at exactly $0$ for all 1{,}025 Playwright sessions (an empty-value signature rather than a distinctive positive value), while bots and humans take positive values. Combined with the \feat{has\_}-indicator channels this makes the agent class linearly separable from two features.}
\label{fig:distributions}
\end{figure}

Figure~\ref{fig:distributions} shows why these features work: agents occupy a behaviorally distinct region from both humans and bots. On \feat{teleport\_click\_ratio} (redefined as ``fraction of clicks preceded within 100\,ms by a raw-event mousemove jump''), all 1{,}025 Playwright-driven agent sessions are at exactly 0 (Playwright emits no such raw jumps at all), while humans cluster around 0.05 and bots spread between 0 and 1. The separation is therefore an \emph{empty-value vs.\ positive-value} signature, made explicit to the classifier through the paired \feat{has\_teleport\_click\_ratio} missingness indicator. The agent distribution is non-overlapping with both other classes through the combination of the feature and its indicator, which is why even a two-feature classifier achieves 100\% agent recall.

\subsection{Key Findings}

\begin{enumerate}
    \item \textbf{Detection is easy under API-driven automation because automation APIs do not emit the raw pointer-move and wheel-delta streams that a physical input device produces.} The signature is an \emph{absence} pattern (zero raw-event teleports before clicks, zero scroll deltas, sparse or empty pre-click mousemove tracks) surfaced through explicit missingness indicators. It is stable across 10 seeds and 3 model families (30 runs, zero agent misclassifications; \S\ref{sec:multiseed}).
    \item \textbf{The detection signal is an API execution artifact, not an agent cognition signal.} Backward elimination removes \feat{teleport\_click\_ratio} at step~9 and all four original ``pause-think-act'' features (\feat{inter\_action\_gap\_variance}, \feat{inter\_action\_gap\_bimodality}, \feat{action\_burst\_ratio}, \feat{time\_to\_first\_action}) by step~10 without breaking agent recall. The surviving 2-feature necessary set \{\feat{mouse\_event\_rate}, \feat{click\_duration\_std}\} contains no cognition-correlated feature; the signal lives in how browser automation APIs generate events.
    \item \textbf{Two features are sufficient for 100\% agent detection with unit agent precision; five features reach macro-F1\,$\geq$\,0.99.} The $k$\,=\,2 best exhaustive subset \{\feat{mouse\_event\_rate}, \feat{teleport\_click\_ratio}\} gives held-out test macro-F1 = 0.926 with agent precision = 0.994 and ROC-AUC = 1.000 at every evasion level; the backward-elimination 2-feature set \{\feat{mouse\_event\_rate}, \feat{click\_duration\_std}\} gives macro-F1 = 0.904 with agent precision = 1.000 and the same ROC. Five features lift macro-F1 to 0.991 (Table~\ref{tab:exhaustive}). On the full 17-feature set, 10-seed macro-F1 95\% CIs are $[0.986, 0.994]$ (SAINT), $[0.993, 0.997]$ (RF), and $[0.993, 0.998]$ (GBM) (Table~\ref{tab:multiseed}).
    \item \textbf{The $k$\,=\,1 regime is a diagnostic trap, not a deployable result.} The one solo-``perfect'' feature (\feat{cursor\_path\_linearity}) achieves 100\% agent recall only by collapsing the classifier to always predict ``agent'' (agent precision = 0.33, macro-F1 = 0.17, ROC-AUC = 0.50). This failure mode is invisible to a recall-only evaluation; our per-level confusion matrices and mixed-set ROC/PR metrics (\S\ref{sec:optimization}) catch it automatically.
    \item \textbf{15/17 features are redundant within our experimental scope.} The first 8 features removed by backward elimination \emph{improve} held-out F1 (from 0.987 to 0.996) by reducing overfitting; features that are redundant here may become necessary on harder distributions (e.g.\ agents operating through alternative frameworks or OS-level input simulation).
    \item \textbf{Detection is not a task artifact.} Removing all task-correlated features (typing speed, scroll speed, keyboard/scroll presence flags, and agent-specific gap features) leaves agent F1 at 1.000 and evasion detection at 100\% across all levels (\S\ref{sec:taskablation}). The signal is in mouse dynamics and event-class presence, not task identity.
\end{enumerate}

\section{Discussion \& Conclusion}
\label{sec:discussion}

\paragraph{What We Are Actually Detecting.}
Our results show that the primary detection signal arises not from high-level agent reasoning behavior but from low-level execution artifacts introduced by browser automation APIs. In this sense, our detector is better understood as identifying \emph{automation pipelines} than as identifying agent cognition. The ``pause-think-act'' hypothesis (\S\ref{sec:pausethinkact}) motivated our agent-specific features, yet none of them appear in the 2-feature backward-elimination necessary set, and all four are removable by step~10 of backward elimination without any drop in agent recall. The signal lives in how automation APIs generate (and fail to generate) events: near-zero raw-mousemove rate, near-zero click-duration variance, absent wheel events. This distinction matters: agents that bypass browser APIs entirely (e.g.\ via OS-level input simulation or hardware mouse emulators) would likely evade our detector. Extending detection to such paradigms is an important direction for future work.

\paragraph{Why So Few Features Suffice.}
The backward-elimination necessary features, \feat{mouse\_event\_rate} and \feat{click\_duration\_std}, capture two complementary properties of the \emph{event stream} generated by browser automation APIs: the overall cadence of raw mousemove emissions (essentially zero-or-sparse for Playwright) and the variability of click hold times (essentially zero because programmatic clicks have flat mouseup-after-mousedown delays). Playwright's \feat{page.mouse.move()} and \feat{page.click()} generate events at programmatic intervals determined by the execution loop, not by the physical dynamics of a human hand. Even when trajectory coordinates are human-derived (Level~5), the event generation pattern (rate, click-duration variance, presence of event classes) bears the signature of API-driven execution. This is a structural rather than incidental signal, which is why two behavioral features suffice to isolate agents at agent-precision = 1.000 and why the signal is redundantly encoded across \feat{mouse\_event\_rate}, \feat{click\_duration\_std}, \feat{teleport\_click\_ratio}, and the missingness-indicator channels.

\paragraph{Implications for Practitioners.}
Within the current dominant deployment setting of API-driven browser agents, our results provide a concrete deployment recommendation: instrument \{\feat{mouse\_event\_rate}, \feat{click\_duration\_std}\} (or the equivalent \{\feat{mouse\_event\_rate}, \feat{teleport\_click\_ratio}\}) for robust agent isolation, and add \feat{typing\_speed\_std}, \feat{mouse\_teleportation}, and \feat{scroll\_speed\_variance} for full 3-class discrimination (macro-F1 $\geq$\,0.99). The remaining 12 features add instrumentation cost without improving detection. Explicitly, \textbf{avoid single-feature deployments}: the one solo-``perfect'' feature (\feat{cursor\_path\_linearity}) we surfaced collapses to an always-agent classifier and is useless in production. This minimal set is robust against all evasion strategies operating at the trajectory and timing level.

\paragraph{Implications for Evasion.}
The finding that two features suffice for unit-precision isolation is paradoxically good news for evasion research: it identifies exactly what must be matched. However, the necessary features are event-stream properties (raw mousemove rate, click-duration variance, wheel-event presence) that cannot be controlled through trajectory manipulation or timing sampling; they arise from the CDP-mediated execution of automation APIs themselves. Because Puppeteer, Playwright, and Selenium-with-CDP share the same underlying protocol~\cite{cloudflare2025browserrun,datadome2023cdp}, we expect the signal to be a structural property of CDP-driven automation rather than a Playwright-specific artifact. Breaking detection would plausibly require abandoning CDP entirely: OS-level input simulation, hardware mouse emulators, DOM-native extension control~\cite{rtrvr2024domnative}, or pixel-level computer-use agents that feed input through the operating system rather than the DevTools Protocol.

\paragraph{Limitations.}
Our experiments are run on Playwright-driven agents, but the underlying signal is a property of the Chrome DevTools Protocol (CDP) control channel rather than Playwright specifically. Industry analyses report that Puppeteer and Playwright are built on CDP and that agent frameworks ``already speak CDP natively''~\cite{cloudflare2025browserrun}, and that CDP-based detection ``targets the underlying technology'' used by modern automation stacks~\cite{datadome2023cdp}. We therefore expect the teleport-click and event-stream signatures we exploit to generalize to Puppeteer, Selenium~4 with CDP bindings, \texttt{chrome.debugger}-based extensions, and raw CDP clients that dispatch input via \texttt{Input.dispatchMouseEvent}. However, we are aware of no published cross-framework measurement study that verifies mouse-event timing distributions are empirically equivalent across these stacks, so this generalization remains a testable prediction rather than a demonstrated result. Execution paths that bypass CDP entirely are outside our scope and would likely require different detection approaches: OS-level input injection (e.g., \texttt{xdotool}, PyAutoGUI, or the macOS Accessibility API used by Anthropic Computer Use~\cite{anthropic2024computeruse}) traverses the full browser input pipeline and does not exhibit CDP-side effects, and DOM-native extension-based architectures such as rtrvr.ai~\cite{rtrvr2024domnative} deliberately avoid CDP to evade this class of detector. WebDriver BiDi / Selenium~4's bidirectional protocol should be treated as uncertain: it is documented as a separate protocol distinct from CDP~\cite{selenium-bidi}, and whether its timing artifacts converge to the CDP detection surface is an open empirical question. Testing these evasion paths (Puppeteer replication, raw CDP, OS-level injection) is the priority for follow-up work. The exhaustive feature search covers $k$\,=\,1\ldots5; the full search space for $k$\,=\,6\ldots16 (52{,}768 additional subsets) would likely show continued growth in perfect-subset density.

\emph{Claim separation.} Our results contain two categories of claims with different evidential strength. \textbf{Bot-independent claims} (the 39.1\%/34.5\% binary detection gap, 100\% three-class agent detection across the full evasion ladder, and the finding that few features suffice for 100\% agent recall) rest on real data for both the human and agent sides of the boundary and are unaffected by bot simulation choices. \textbf{Bot-dependent claims} (macro-F1 values and the minimal feature set for 3-class discrimination) depend on our synthetic bot class (three archetypes sampled from literature-grounded Gaussian distributions). While the agent detection results are robust, the specific macro-F1 confidence intervals ($[0.986, 0.998]$ across SAINT/GBM/RF) and the exact minimal feature set reflect our simulator assumptions and may shift under different bot distributions.

Our feature subset optimization uses GradientBoosting rather than SAINT for computational tractability in the exhaustive search (9{,}401 GBMs over $\binom{17}{k}$ for $k$\,=\,1\ldots5); we re-trained SAINT on the $k$\,=\,2 best exhaustive subset, the 2-feature backward-elimination necessary set, and the $k$\,=\,5 best subset (Table~\ref{tab:saint_subset}). SAINT preserves agent recall = 1.000 at every evasion level on all three subsets; macro-F1 trails the GBM by 0.01--0.03 on the smallest subsets, an expected gap given SAINT's higher parameter count and our lighter validation training budget (30 epochs with patience~5, vs.\ the 50-epoch main protocol in Table~\ref{tab:hyperparams}). Our agent data covers 10 web tasks driven by a single LLM (Claude) through Playwright and MCP; other LLMs, automation frameworks (Selenium, Puppeteer), and task distributions may produce different behavioral profiles. Human data is sourced from CAPTCHA-solving sessions, which differ in task structure from agent web tasks; our task-domain ablation (\S\ref{sec:taskablation}) shows this does not affect agent detection, though future work should validate on matched-task human data. The replay evaluation set is now $n$\,=\,524 sessions (all sessions with non-empty event streams in \texttt{data/raw/replay\_evasion\_sessions}; the previous count of 165 reflected an earlier snapshot of the same directory, and both replay and feature-optimization scripts now share one loader so counts cannot drift again).

\paragraph{Ethical Considerations.}
All human behavioral data is sourced from CaptchaSolve30k~\cite{captchasolve30k}, a publicly available dataset. Our detection research focuses on fundamental API-level signals rather than novel evasion techniques; the evasion methods we evaluate (GAN trajectories, human replay) are well-established in the adversarial ML literature~\cite{acien2021becaptcha,goodfellow2014generative}. While detection techniques could theoretically inform evasion, our findings suggest that the fundamental signal (API-driven event generation) cannot be circumvented without abandoning browser automation APIs entirely, limiting the dual-use risk.

\paragraph{Reproducibility.}
All experiments use 70/15/15 stratified train/val/test splits. Key results are validated across 10 random seeds ($s$\,=\,42\ldots51) with 95\% confidence intervals reported via the $t$-distribution (\S\ref{sec:multiseed}). The complete codebase, trained models, and dataset will be made available upon publication.

\paragraph{Conclusion.}
We present the first feature-level analysis of what makes behavioral agent detection work within Playwright-mediated browser automation. Through exhaustive search over 9{,}401 feature subsets, forward selection, and backward elimination, we show that \textbf{two} behavioral features suffice for 100\% agent recall with unit agent precision at every evasion level: either \{\feat{mouse\_event\_rate}, \feat{teleport\_click\_ratio}\} from the exhaustive search (held-out macro-F1 = 0.926) or \{\feat{mouse\_event\_rate}, \feat{click\_duration\_std}\} from backward elimination (macro-F1 = 0.904). Five features are enough to reach macro-F1 $\geq$\,0.99 for full three-class discrimination. The $k$\,=\,1 regime is degenerate and should not be reported as a deployable result: the one solo-``perfect'' indicator we surfaced collapses the classifier to always predict ``agent'' (agent precision = 0.33, ROC-AUC = 0.5). On the full 17-feature set, 10-seed macro-F1 95\% CIs are $[0.986, 0.994]$ (SAINT), $[0.993, 0.997]$ (RF), and $[0.993, 0.998]$ (GBM). The 12--15 features outside the minimal sets are redundant \emph{within our experimental scope} and may prove necessary as agents diversify across frameworks, LLMs, and evasion strategies. A task-domain ablation confirms that the signal resides in event-stream structure, not in task-structural differences between data sources. Detection succeeds because automation APIs do not emit the raw pointer-move and wheel-delta streams that a physical input device produces; the resulting absence signature survives trajectory manipulation. Because Playwright, Puppeteer, and Selenium-with-CDP share the Chrome DevTools Protocol as a common substrate~\cite{cloudflare2025browserrun,datadome2023cdp}, we expect the signal to generalize across CDP-mediated automation, though empirical verification on non-Playwright stacks is left to future work. The signal is robust, minimal, and actionable within our threat model, and degrades gracefully when any single feature is removed.


\bibliographystyle{ACM-Reference-Format}


\clearpage
\appendix

\section{Dataset Processing Details}
\label{app:data}

\paragraph{Event Parsing.} The raw data uses two schema formats: (1)~\texttt{tickInputs} arrays containing \{$x$, $y$, \texttt{isDown}, \texttt{sampleIndex}\} objects, and (2)~flat arrays of $x$/$y$/timestamp/button\_state values. We implement a unified parser that handles both formats and converts tick indices to millisecond timestamps via $t_{\text{ms}} = i \cdot \frac{d}{n_{\text{ticks}}}$, where $d$ is the session duration and $n_{\text{ticks}}$ is the total tick count (default 4.17\,ms/tick for 240\,Hz sampling).

\paragraph{Button State Machine.} Rather than treating raw button states as instantaneous signals, we implement a state machine that tracks the previous button state and emits \texttt{mousedown}/\texttt{mouseup} transitions. This correctly captures click edges (the onset and release of a mouse button press), which are critical for computing click duration features.

\paragraph{Trajectory Segmentation.} We extract cursor movement segments bounded by action events (click, keydown, submit), requiring a minimum of 5 raw points per segment. Each segment is represented in cumulative offset coordinates: $\mathbf{p}_i = (x_i - x_0, y_i - y_0, t_i - t_0)$, which normalizes for absolute screen position and enables comparison across sessions.

\paragraph{Resampling.} All trajectories are resampled to 60\,Hz ($\Delta t = 16.67$\,ms) via linear interpolation and normalized to 25 timesteps using \texttt{numpy.interp}. This produces fixed-length trajectory representations suitable for both feature extraction and GAN training.

\section{Feature Definitions}
\label{app:features}

\subsection{Trajectory Features}

\paragraph{Trajectory Entropy.} We quantize cursor direction angles into $K$\,=\,20 equal-width bins and compute Shannon entropy:
\begin{equation}
    H_{\text{traj}} = -\sum_{k=1}^{K} p_k \log_2 p_k
\end{equation}
where $p_k$ is the fraction of direction changes falling in bin $k$. Higher entropy indicates more varied movement directions, characteristic of human cursor behavior~\cite{pusara2004mouse}.

\paragraph{Movement Efficiency.} The ratio of straight-line displacement to total path length:
\begin{equation}
    \eta = \frac{\|\mathbf{p}_N - \mathbf{p}_0\|}{\sum_{i=1}^{N}\|\mathbf{p}_i - \mathbf{p}_{i-1}\|}
\end{equation}
Values near 1.0 indicate direct paths (typical of automated movement); lower values indicate the indirect, corrective paths characteristic of human motor control~\cite{ahmed2007mouse}.

\paragraph{Mouse Teleportation.} The count of instantaneous cursor jumps exceeding a threshold:
\begin{equation}
    \tau = \bigl|\{i : \|\mathbf{p}_i - \mathbf{p}_{i-1}\| > \delta_{\text{tp}}\}\bigr|, \quad \delta_{\text{tp}} = 100\text{\,px}
\end{equation}
Teleportation occurs when browser automation APIs set cursor position directly rather than generating intermediate movement events.

\paragraph{Cursor Path Linearity.} Average linearity over sliding windows of $w$\,=\,10 points:
\begin{equation}
    \ell = \frac{1}{|\mathcal{W}|}\sum_{(a,b) \in \mathcal{W}} \frac{\|\mathbf{p}_b - \mathbf{p}_a\|}{\sum_{i=a+1}^{b}\|\mathbf{p}_i - \mathbf{p}_{i-1}\|}
\end{equation}
where $\mathcal{W}$ is the set of all consecutive $w$-point windows.

\subsection{Click Features}

\paragraph{Click Duration Mean and Std.} Statistics of the time between matched \texttt{mousedown} and \texttt{mouseup} events:
\begin{equation}
    \bar{d}_{\text{click}} = \frac{1}{n_c}\sum_{j=1}^{n_c}(t^{\text{up}}_j - t^{\text{down}}_j), \quad \sigma_{\text{click}} = \text{std}(\{t^{\text{up}}_j - t^{\text{down}}_j\}_{j=1}^{n_c})
\end{equation}
Automation APIs produce unnaturally fast ($<$10\,ms) and uniform click durations, while human clicks show wider variance ($\sigma \approx 30$\,ms) centered around 100--120\,ms~\cite{mondal2017continuous}.

\subsection{Timing Features}

\paragraph{Mouse Event Rate.} The frequency of \texttt{mousemove} events: $r = |\text{mousemove}| / T_{\text{session}}$, measured in Hz. Human interaction produces dense event streams (10--50\,Hz), while automation APIs generate sparse, programmatic events~\cite{chu2018bot}.

\paragraph{Inter-Event Entropy.} Shannon entropy of the distribution of inter-event time gaps across all event types:
\begin{equation}
    H_{\text{evt}} = -\sum_{k=1}^{K} p_k \log_2 p_k
\end{equation}
computed over $K$\,=\,20 bins of inter-event gaps. Higher entropy indicates more varied timing, characteristic of human interaction.

\paragraph{Inter-Action Gap Variance and Bimodality.} The variance of gaps between action events (click, keydown, submit) captures timing regularity. Bimodality is measured as the fraction of gaps exceeding a burst threshold:
\begin{equation}
    \beta = \frac{|\{\Delta t_i > \delta_{\text{burst}}\}|}{|\{\Delta t_i\}|}, \quad \delta_{\text{burst}} = 1000\text{\,ms}
\end{equation}

\paragraph{Action Burst Ratio.} The complement of bimodality: the fraction of inter-action gaps below the burst threshold:
\begin{equation}
    r_{\text{burst}} = \frac{|\{\Delta t_i < \delta_{\text{burst}}\}|}{|\{\Delta t_i\}|}
\end{equation}
Agents exhibit high burst ratios due to rapid-fire actions during the ``act'' phase of their pause-think-act cycle.

\paragraph{Time to First Action.} The delay $t_{\text{first\_action}} - t_{\text{first\_event}}$ captures the ``page read time'' before the first interaction.

\paragraph{Typing Speed Mean and Std.} Statistics of inter-keydown intervals. Human typing shows high variance ($\sigma \approx 55$\,ms) reflecting natural motor variability, while automation produces uniform timing.

\subsection{Agent-Specific Features}

\paragraph{Teleport-Click Ratio.} The fraction of clicks preceded by cursor teleportation:
\begin{equation}
    r_{\text{tc}} = \min\!\left(1, \frac{\tau}{n_{\text{clicks}}}\right)
\end{equation}
This is the single most discriminative feature in our analysis (\S\ref{sec:optimization}), as automation APIs typically move the cursor to the target element and immediately click, producing a characteristic teleport-then-click pattern.

\paragraph{Scroll Speed Variance.} The variance of inter-scroll-event time gaps: $\text{Var}(\{\Delta t_{\text{scroll}}\})$. Programmatic scrolling produces uniform intervals, while human scrolling is irregular.

\subsection{Feature Normalization}

All continuous features are min-max normalized to $[0, 1]$:
\begin{equation}
    \tilde{x}_i = \frac{x_i - x_{\min}}{x_{\max} - x_{\min}}
\end{equation}
with $x_{\min}$ and $x_{\max}$ fitted on the balanced training set only. For nullable features, we add binary indicator columns $h_i = \mathbb{1}[x_i \neq \text{null}]$. Binary features are directly converted to \texttt{float32}.

\section{SAINT Architecture and Training Details}
\label{app:model}

\subsection{Per-Feature Embedding}

Unlike vision or language tasks where inputs share a common modality, tabular features are heterogeneous: each feature has its own scale, distribution, and semantics. Following SAINT~\cite{somepalli2021saint} and FT-Transformer~\cite{gorishniy2021revisiting}, we project each feature independently into a shared embedding space.

For each continuous feature $x_i$ ($i \in \mathcal{C}$), we learn a separate linear projection:
\begin{equation}
    \mathbf{e}_i = \mathbf{W}_i x_i + \mathbf{b}_i \in \mathbb{R}^{d}
\end{equation}
where $\mathbf{W}_i \in \mathbb{R}^{d \times 1}$ and $\mathbf{b}_i \in \mathbb{R}^{d}$. For each binary feature $b_j$ ($j \in \mathcal{B}$), we use a learned embedding lookup:
\begin{equation}
    \mathbf{e}_j = \texttt{Embed}_j(b_j) \in \mathbb{R}^{d}
\end{equation}
where \texttt{Embed}$_j$ is a $2 \times d$ embedding table. This per-feature approach allows each feature to learn its own representation, in contrast to TabTransformer~\cite{huang2020tabtransformer} which only applies attention to categorical features. We use $d$\,=\,32 throughout.

\subsection{CLS Token and Positional Encoding}

A learnable CLS token $\mathbf{c} \in \mathbb{R}^{d}$ is prepended to the feature embedding sequence, following the BERT~\cite{devlin2019bert} convention for aggregating sequence-level representations. Learnable positional encodings $\mathbf{P} \in \mathbb{R}^{(n+1) \times d}$ are added:
\begin{equation}
    \mathbf{S}^{(0)} = [\mathbf{c};\, \mathbf{e}_1;\, \ldots;\, \mathbf{e}_n] + \mathbf{P}
\end{equation}
We use learnable rather than sinusoidal positional encodings because tabular features have no inherent sequential ordering; the positional encoding must learn which features interact most strongly, rather than encoding proximity.

\subsection{Transformer Blocks}

The embedded sequence passes through $L$\,=\,2 transformer blocks. Each block applies multi-head self-attention~\cite{vaswani2017attention} followed by a position-wise feed-forward network (FFN), with residual connections and layer normalization:

\paragraph{Multi-Head Self-Attention.} With $h$\,=\,4 attention heads:
\begin{align}
    \text{Attn}(\mathbf{Q}, \mathbf{K}, \mathbf{V}) &= \text{softmax}\!\left(\frac{\mathbf{Q}\mathbf{K}^\top}{\sqrt{d_k}}\right)\mathbf{V} \\
    \text{head}_i &= \text{Attn}(\mathbf{S}\mathbf{W}_i^Q, \mathbf{S}\mathbf{W}_i^K, \mathbf{S}\mathbf{W}_i^V) \\
    \text{MHA}(\mathbf{S}) &= [\text{head}_1; \ldots; \text{head}_h]\mathbf{W}^O
\end{align}
where $d_k = d/h = 8$. Self-attention allows each feature to attend to every other feature, learning cross-feature interactions that are critical for tabular data~\cite{somepalli2021saint}.

\paragraph{Feed-Forward Network.} A two-layer FFN with GELU activation~\cite{hendrycks2016gelu} and expansion factor 2:
\begin{equation}
    \text{FFN}(\mathbf{x}) = \mathbf{W}_2 \cdot \text{GELU}(\mathbf{W}_1 \mathbf{x} + \mathbf{b}_1) + \mathbf{b}_2
\end{equation}
where $\mathbf{W}_1 \in \mathbb{R}^{2d \times d}$ and $\mathbf{W}_2 \in \mathbb{R}^{d \times 2d}$.

\paragraph{Residual Connections.} Each sub-layer uses post-norm residual connections:
\begin{align}
    \mathbf{S}' &= \text{LayerNorm}(\mathbf{S} + \text{MHA}(\mathbf{S})) \\
    \mathbf{S}^{(\ell+1)} &= \text{LayerNorm}(\mathbf{S}' + \text{FFN}(\mathbf{S}'))
\end{align}

\paragraph{Why 2 Blocks Suffice.} Unlike NLP or vision tasks where deep stacks capture hierarchical abstractions over long sequences, tabular data has a fixed, small number of features ($n$\,=\,16--23). Two attention layers allow each feature to attend to all others directly (layer~1) and to attend to those attention-informed representations (layer~2), which suffices for capturing pairwise and higher-order feature interactions. Empirically, adding more blocks yielded no improvement on our dataset.

\subsection{Denoising Pre-Training}
\label{sec:denoising}

Following SAINT~\cite{somepalli2021saint}, we pre-train the encoder using a denoising objective. For each sample, we randomly mask 30\% of input features by replacing them with zeros. The encoder processes the corrupted input, and per-feature reconstruction heads predict the original values from the encoded representations.

For continuous features, a linear head predicts the original value:
\begin{equation}
    \hat{x}_i = \mathbf{w}_i^\top \mathbf{s}_i + c_i
\end{equation}
For binary features, a linear head followed by sigmoid predicts the original label:
\begin{equation}
    \hat{b}_j = \sigma(\mathbf{w}_j^\top \mathbf{s}_j + c_j)
\end{equation}
where $\mathbf{s}_i$ is the encoder's output for the $i$-th feature token (CLS token excluded).

The denoising loss is computed only over masked positions:
\begin{equation}
    \mathcal{L}_{\text{denoise}} = \frac{1}{|\mathcal{M}_c|}\sum_{i \in \mathcal{M}_c} (x_i - \hat{x}_i)^2 + \frac{1}{|\mathcal{M}_b|}\sum_{j \in \mathcal{M}_b} \text{BCE}(b_j, \hat{b}_j)
\end{equation}
where $\mathcal{M}_c$ and $\mathcal{M}_b$ are the sets of masked continuous and binary features, respectively. Pre-training uses a separate Adam optimizer and runs for 50 epochs before supervised fine-tuning begins.

We choose denoising over contrastive pre-training because our dataset is small ($\sim$6{,}000 samples): contrastive methods require diverse positive pairs to learn useful representations, while denoising directly learns feature dependencies from individual samples. This aligns with findings that denoising pre-training is more effective than contrastive objectives for small tabular datasets~\cite{somepalli2021saint}.

\subsection{Binary Classification Head}

For the binary (human-vs-bot) detection task, the CLS token output feeds a two-layer classification head:
\begin{equation}
    \hat{y} = \sigma(\mathbf{w}_2^\top \text{ReLU}(\mathbf{W}_1 \mathbf{s}_{\text{CLS}} + \mathbf{b}_1) + b_2)
\end{equation}
where $\mathbf{W}_1 \in \mathbb{R}^{d \times d}$. Training uses binary cross-entropy with logits (BCEWithLogitsLoss). We use CLS token pooling rather than mean pooling over all feature tokens because the CLS token learns to aggregate information during attention, acting as a global summary~\cite{devlin2019bert}.

\subsection{Three-Class Adaptation}

For three-class detection (human/bot/agent), we replace the binary head with a multi-class head:
\begin{equation}
    \hat{\mathbf{y}} = \mathbf{W}_2 \cdot \text{Dropout}(\text{ReLU}(\mathbf{W}_1 \mathbf{s}_{\text{CLS}} + \mathbf{b}_1)) + \mathbf{b}_2 \in \mathbb{R}^{3}
\end{equation}
with Dropout($p$\,=\,0.1) for regularization. Training uses cross-entropy loss with optional class weighting to handle any residual class imbalance after downsampling.

The key insight is that the \emph{same encoder architecture} supports both tasks: only the classification head changes. Yet the three-class model achieves dramatically different detection capability: perfect agent detection (F1\,=\,1.000) where the binary SAINT model allows 34.5\% evasion (MLP 39.1\%).

\paragraph{Expected Calibration Error.} We evaluate prediction calibration using ECE~\cite{naeini2015ece} with $M$\,=\,10 confidence bins:
\begin{equation}
    \text{ECE} = \sum_{m=1}^{M} \frac{|B_m|}{n} |\text{acc}(B_m) - \text{conf}(B_m)|
\end{equation}
where $B_m$ is the set of samples whose maximum predicted probability falls in bin $m$, $\text{acc}(B_m)$ is the accuracy within that bin, and $\text{conf}(B_m)$ is the mean predicted confidence.

\subsection{Training Procedure}

Supervised fine-tuning follows the denoising pre-training phase described in \S\ref{sec:denoising}.

\paragraph{Optimizer.} Adam~\cite{loshchilov2019adamw} with learning rate $\eta = 10^{-3}$ and weight decay $\lambda = 10^{-4}$.

\paragraph{Learning Rate Schedule.} Cosine annealing~\cite{loshchilov2019adamw}:
\begin{equation}
    \eta_t = \eta_{\min} + \frac{1}{2}(\eta_{\max} - \eta_{\min})\left(1 + \cos\frac{t\pi}{T}\right)
\end{equation}
where $T$ is the total number of epochs.

\paragraph{Early Stopping.} We monitor validation macro-F1 with patience of 15 epochs. The best model checkpoint (highest validation macro-F1) is restored after training completes.

Table~\ref{tab:hyperparams} summarizes all hyperparameters.

\begin{table}[t]
\centering
\caption{Hyperparameters for SAINT training.}
\label{tab:hyperparams}
\small
\begin{tabular}{lc}
\toprule
Hyperparameter & Value \\
\midrule
Embedding dimension $d$ & 32 \\
Attention heads $h$ & 4 \\
Transformer blocks $L$ & 2 \\
FFN expansion factor & 2 \\
Dropout (attention) & 0.1 \\
Dropout (classification head) & 0.1 \\
Denoising mask rate & 30\% \\
Pre-training epochs & 50 \\
Learning rate $\eta$ & $10^{-3}$ \\
Weight decay $\lambda$ & $10^{-4}$ \\
Scheduler & Cosine annealing \\
Early stopping patience & 15 epochs \\
Batch size & 64 \\
Random seed & 42 \\
\bottomrule
\end{tabular}
\end{table}


\begin{figure*}[t]
\centering
\begin{tikzpicture}[
    node distance=0.6cm and 0.8cm,
    box/.style={draw, rounded corners, minimum height=0.7cm, minimum width=1.4cm, font=\small, fill=#1},
    box/.default=blue!8,
    arrow/.style={-Stealth, thick},
    label/.style={font=\scriptsize, text=gray!70!black},
    every node/.style={align=center},
]

\node[box=orange!12] (x1) {$x_1$};
\node[box=orange!12, right=0.3cm of x1] (x2) {$x_2$};
\node[right=0.3cm of x2] (dots1) {$\cdots$};
\node[box=orange!12, right=0.3cm of dots1] (xn) {$x_n$};
\node[box=green!12, right=0.3cm of xn] (b1) {$b_1$};
\node[right=0.3cm of b1] (dots2) {$\cdots$};
\node[box=green!12, right=0.3cm of dots2] (bm) {$b_m$};

\node[label, above=0.15cm of dots1] {continuous};
\node[label, above=0.15cm of dots2] {binary};

\node[box=blue!15, below=0.8cm of x1] (e1) {$\mathbf{W}_1 x_1{+}\mathbf{b}_1$};
\node[box=blue!15, below=0.8cm of x2] (e2) {$\mathbf{W}_2 x_2{+}\mathbf{b}_2$};
\node[below=0.8cm of dots1] (edots) {$\cdots$};
\node[box=blue!15, below=0.8cm of xn] (en) {$\mathbf{W}_n x_n{+}\mathbf{b}_n$};
\node[box=green!15, below=0.8cm of b1] (eb1) {$\text{Emb}(b_1)$};
\node[below=0.8cm of dots2] (ebdots) {$\cdots$};
\node[box=green!15, below=0.8cm of bm] (ebm) {$\text{Emb}(b_m)$};

\node[box=red!15, left=0.5cm of e1] (cls) {\textbf{CLS}};

\foreach \s/\t in {x1/e1, x2/e2, xn/en, b1/eb1, bm/ebm}
    \draw[arrow] (\s) -- (\t);

\node[box=purple!10, below=0.8cm of edots, minimum width=10cm] (pos) {$+ \;\mathbf{P} \in \mathbb{R}^{(n+1) \times d}$ \quad (learnable positional encoding)};

\draw[arrow] (cls) -- (cls |- pos.north);
\draw[arrow] (e1) -- (e1 |- pos.north);
\draw[arrow] (e2) -- (e2 |- pos.north);
\draw[arrow] (en) -- (en |- pos.north);
\draw[arrow] (eb1) -- (eb1 |- pos.north);
\draw[arrow] (ebm) -- (ebm |- pos.north);

\node[box=yellow!15, below=0.8cm of pos, minimum width=10cm, minimum height=0.9cm] (tf1) {\textbf{Transformer Block 1:} MHA($h$=4, $d_k$=8) $\to$ LayerNorm $\to$ FFN(GELU, $d{\to}2d{\to}d$) $\to$ LayerNorm};

\node[box=yellow!15, below=0.5cm of tf1, minimum width=10cm, minimum height=0.9cm] (tf2) {\textbf{Transformer Block 2:} MHA($h$=4, $d_k$=8) $\to$ LayerNorm $\to$ FFN(GELU, $d{\to}2d{\to}d$) $\to$ LayerNorm};

\draw[arrow] (pos) -- (tf1);
\draw[arrow] (tf1) -- (tf2);

\node[box=red!15, below left=0.8cm and -0.5cm of tf2] (clsout) {CLS output\\$\mathbf{s}_{\text{CLS}} \in \mathbb{R}^d$};

\node[box=gray!15, below right=0.8cm and -0.5cm of tf2] (featout) {Feature tokens\\$\mathbf{s}_1, \ldots, \mathbf{s}_n$};

\draw[arrow] (tf2.south) -- ++(0,-0.3) -| (clsout.north);
\draw[arrow] (tf2.south) -- ++(0,-0.3) -| (featout.north);

\node[box=blue!20, below left=0.8cm and 0.5cm of clsout] (binary) {Binary Head\\$\sigma(\mathbf{W}\mathbf{s} + \mathbf{b})$};
\node[box=blue!20, below right=0.8cm and 0.5cm of clsout] (multi) {3-Class Head\\softmax($\mathbf{W}\mathbf{s} + \mathbf{b}$)};

\draw[arrow] (clsout) -- (binary);
\draw[arrow] (clsout) -- (multi);

\node[box=gray!20, below=0.8cm of featout] (denoise) {Denoising Head\\(pre-training only)};
\draw[arrow, dashed] (featout) -- (denoise);

\node[below=0.3cm of binary, font=\small] {$P(\text{bot})$};
\node[below=0.3cm of multi, font=\small] {$P(\text{H}), P(\text{B}), P(\text{A})$};
\node[below=0.3cm of denoise, font=\small, text=gray] {$\hat{x}_1, \ldots, \hat{x}_n$};

\end{tikzpicture}
\caption{SAINT architecture for behavioral agent detection. Input features are independently embedded (continuous via linear projection, binary via embedding lookup), combined with a learnable CLS token and positional encoding, and processed by two transformer blocks. The CLS output feeds either a binary or three-class classification head. During pre-training, feature tokens feed a denoising reconstruction head (dashed).}
\label{fig:architecture}
\end{figure*}
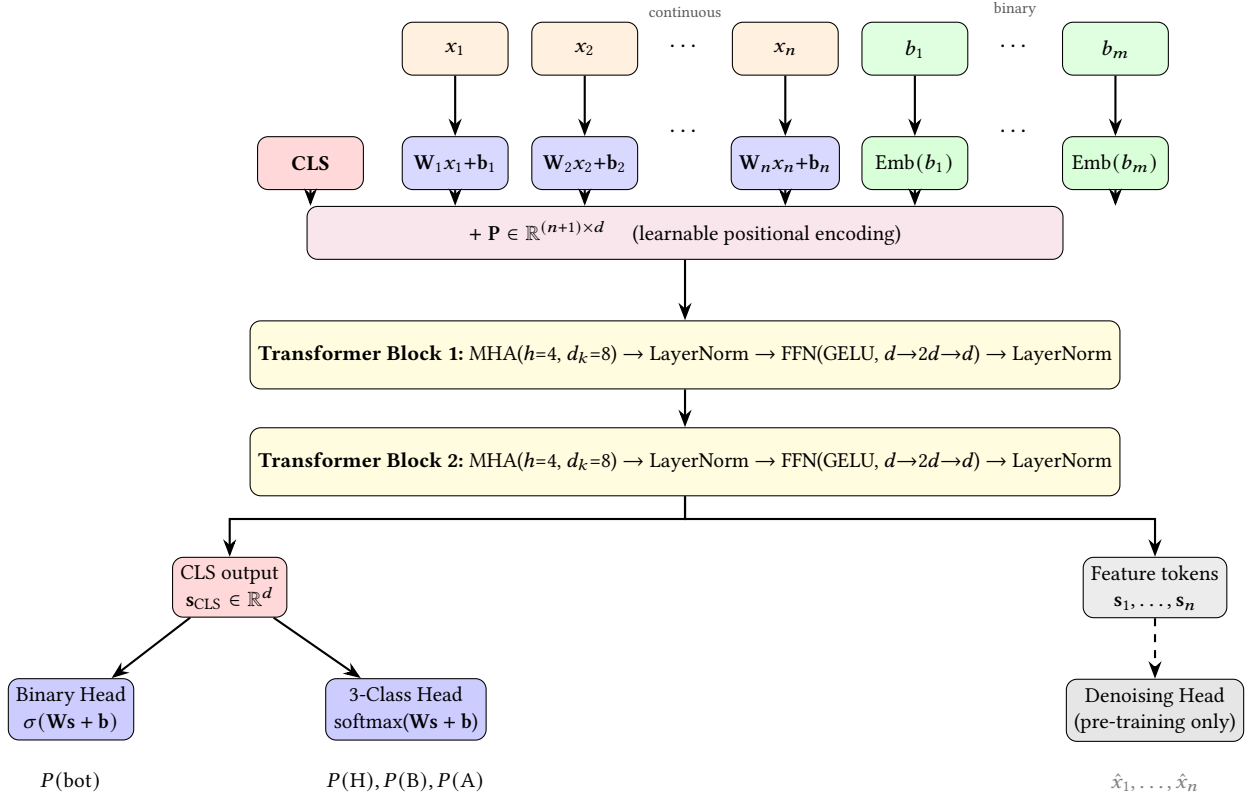

\section{Evasion Methodology}
\label{sec:evasion}

We construct a five-level evasion ladder of escalating attacker sophistication. Level~1 (no evasion) uses unmodified agent behavior. Levels~2--5 apply progressively stronger evasion techniques to test whether detection relies on superficial or fundamental signals.

\subsection{Rule-Based Evasion (Level~2)}
\label{sec:rulebased}

Rule-based evasion applies hand-crafted humanization to the agent's browser interactions. We implement three sub-levels of increasing sophistication:

\paragraph{Mild.} Random inter-action delays sampled from $U(500, 2000)$\,ms. Sequential key presses with random per-character delays in $[80, 200]$\,ms. No cursor movement simulation.

\paragraph{Moderate.} In addition to timing delays, we simulate human-like cursor movement before each click using quadratic B\'{e}zier interpolation. Given start point $\mathbf{p}_0$ and target $\mathbf{p}_2$, we compute a control point $\mathbf{p}_1$ with random offsets ($\pm 50$\,px horizontal, $\pm 30$\,px vertical) and interpolate:
\begin{equation}
    \mathbf{B}(t) = (1-t)^2\mathbf{p}_0 + 2(1-t)t\,\mathbf{p}_1 + t^2\mathbf{p}_2, \quad t \in [0,1]
\end{equation}
sampled at 15 steps with 8--25\,ms per step, producing a 40--125\,Hz movement rate. Gaussian micro-jitter $\mathcal{N}(0, 1.5)$\,px is added per step.

\paragraph{Aggressive.} Additional humanization: pre-task scrolling ($\pm 50$--150\,px), curved cursor paths (20--25 B\'{e}zier steps with stronger control point offsets), typing rhythm variation (base delay 130--140\,ms, variation factor 0.55--0.6), idle micro-movements (3--8 resting hand gestures with Gaussian drift), and page read simulation between major actions.

\subsection{Conditional WGAN-GP (Levels~3--4)}
\label{sec:wgangp}

For GAN-based evasion, we train a conditional Wasserstein GAN with gradient penalty (WGAN-GP)~\cite{arjovsky2017wasserstein,gulrajani2017improved} on 14{,}342 real human trajectory segments from CaptchaSolve30k~\cite{captchasolve30k}.

\paragraph{Generator.} $G(\mathbf{z}, \mathbf{c}) \to \mathbb{R}^{T \times 3}$ maps noise $\mathbf{z} \sim \mathcal{N}(\mathbf{0}, \mathbf{I}_{64})$ and condition $\mathbf{c} = (d, \theta, T_{\text{dur}})$ (distance, angle, duration) to a trajectory of $T$\,=\,25 displacement steps $(\Delta x, \Delta y, \Delta t)$:
\begin{equation}
    G: \mathbb{R}^{64} \times \mathbb{R}^{3} \xrightarrow{\text{FC-BN-ReLU}} \mathbb{R}^{512} \xrightarrow{\text{FC-BN-ReLU}} \mathbb{R}^{75} \xrightarrow{\tanh} \mathbb{R}^{25 \times 3}
\end{equation}
Output displacements are scaled by distance ($\Delta x, \Delta y$) and duration ($\Delta t$) from the condition vector.

\paragraph{Discriminator.} $D(\mathbf{x}, \mathbf{c}) \to \mathbb{R}$ scores trajectories (higher = more realistic), with no sigmoid, consistent with the Wasserstein formulation:
\begin{equation}
    D: \mathbb{R}^{78} \xrightarrow{\text{FC-LeakyReLU-Drop}} \mathbb{R}^{512} \xrightarrow{\text{FC-LeakyReLU-Drop}} \mathbb{R}^{256} \xrightarrow{\text{FC}} \mathbb{R}
\end{equation}
with LeakyReLU($\alpha$\,=\,0.2) and Dropout($p$\,=\,0.3).

\paragraph{Training Objective.} The discriminator minimizes:
\begin{equation}
    \mathcal{L}_D = \underset{\mathbf{z}}{\mathbb{E}}[D(G(\mathbf{z}, \mathbf{c}), \mathbf{c})] - \underset{\mathbf{x}}{\mathbb{E}}[D(\mathbf{x}, \mathbf{c})] + \lambda_{\text{GP}} \underset{\hat{\mathbf{x}}}{\mathbb{E}}\!\left[(\|\nabla_{\hat{\mathbf{x}}} D(\hat{\mathbf{x}}, \mathbf{c})\|_2 - 1)^2\right]
\end{equation}
where $\hat{\mathbf{x}} = \alpha \mathbf{x} + (1-\alpha) G(\mathbf{z}, \mathbf{c})$ with $\alpha \sim U(0,1)$, and $\lambda_{\text{GP}} = 10$.

\paragraph{Smoothness Regularization.} We add an acceleration penalty to encourage smooth trajectories:
\begin{equation}
    \mathcal{L}_{\text{smooth}} = \mathbb{E}\!\left[\frac{1}{T-2}\sum_{t=1}^{T-2}\|\mathbf{v}_{t+1} - \mathbf{v}_t\|^2\right]
\end{equation}
where $\mathbf{v}_t = \mathbf{x}_{t+1} - \mathbf{x}_t$ is the velocity at step $t$. The generator minimizes $\mathcal{L}_G = -\mathbb{E}[D(G(\mathbf{z}, \mathbf{c}), \mathbf{c})] + \lambda_s \mathcal{L}_{\text{smooth}}$ with $\lambda_s = 0.1$.

\paragraph{Fitts's Law Conditioning.} Movement duration is estimated using Fitts's law~\cite{fitts1954information}:
\begin{equation}
    T_{\text{dur}} = a + b \log_2\!\left(\frac{d}{w} + 1\right) + \epsilon
\end{equation}
with $a = 50$\,ms, $b = 150$\,ms, and noise $\epsilon \sim \mathcal{N}(0, 0.15 T_{\text{dur}})$. This grounds trajectory durations in the well-established relationship between movement distance and time~\cite{fitts1954information}.

Training uses $n_{\text{critic}} = 5$ discriminator updates per generator update, Adam($\beta_1\!=\!0.5, \beta_2\!=\!0.9$), $\eta = 10^{-4}$, for 200 epochs with batch size 128.

\paragraph{Level~4 (GAN+).} GAN trajectories are further augmented with Gaussian micro-jitter ($\sigma$\,=\,1.5\,px), Fitts-law timing, and click/gap distributions sampled from human data.

\subsection{Human Trajectory Replay (Level~5)}
\label{sec:replay}

The strongest evasion strategy replays real human trajectories, which by definition match the human feature distribution for trajectory-level features.

\paragraph{Trajectory Index.} We build a KD-tree~\cite{bentley1975kdtree} over $(d, \theta)$ from CaptchaSolve30k sessions, with distance normalized by the median to equalize scales. To avoid distributional overlap between training data and replay trajectories, we split CaptchaSolve30k 80/20: the 80\% training partition is used for human feature extraction and model training, while the 20\% held-out partition is reserved exclusively for Level~5 replay trajectory sourcing. For each required cursor movement, we query the $k$\,=\,5 nearest neighbors and randomly select one, trading optimality for trajectory diversity.

\paragraph{Affine Warping.} A retrieved trajectory with endpoint $(d_s, \theta_s)$ is warped to match the target $(d_t, \theta_t)$ via scaling and rotation:
\begin{equation}
    \begin{pmatrix} x' \\ y' \end{pmatrix} = s \cdot \mathbf{R}_{\Delta\theta} \begin{pmatrix} x \\ y \end{pmatrix}, \quad s = \frac{d_t}{d_s}, \quad \Delta\theta = \theta_t - \theta_s
\end{equation}
where $\mathbf{R}_{\Delta\theta}$ is the 2D rotation matrix. This preserves the trajectory's curvature profile while fitting exact endpoints.

\paragraph{Timing Replay.} Original human $\Delta t$ values are preserved from the source trajectory. Micro-jitter $\mathcal{N}(0, 1.0)$\,px is added at each step. Inter-action gaps are sampled from $\ln\mathcal{N}(6.0, 0.5)$ clipped to $[100, 5000]$\,ms (median $\approx 403$\,ms). Click hold durations follow $\mathcal{N}(110, 30)$\,ms, and typing inter-key gaps follow $\mathcal{N}(180, 60)$\,ms, all matching human distributions.

\paragraph{LLM Decoupling.} Critically, we decouple LLM inference from action execution. Rather than pausing between actions to wait for model output (which would introduce the pause-think-act pattern), we pre-compute all actions and execute them with human-like timing. This eliminates the most obvious agent signal and represents the theoretical strongest evasion for API-based automation.

\end{document}